\documentclass[lettersize,journal]{IEEEtran}
\usepackage{amsmath,amsfonts}
\usepackage{algorithmic}
\usepackage{algorithm}
\usepackage{array}
\usepackage{textcomp}
\usepackage{stfloats}
\usepackage{url}
\usepackage{verbatim}
\usepackage{graphicx}
\usepackage{cite}
\usepackage{subfigure}
\usepackage{url}
\usepackage{multirow}
\usepackage{enumitem}
\setlist[enumerate]{leftmargin=0.9cm, label=(\arabic*)}
\usepackage{arydshln}
\usepackage{amsfonts}
\usepackage{amsmath}
\usepackage{amsthm}

\theoremstyle{definition}

\usepackage{booktabs}

\begin{document}

\title{Aligning Multiple Knowledge Graphs in \\ a Single Pass}

\author{Yaming Yang, Zhe Wang, Ziyu Guan$^{*}$, Wei Zhao, Weigang Lu, Xinyan Huang, Jiangtao Cui, Xiaofei He%
\thanks{* Corresponding author}%
\thanks{Y. Yang, Z. Wang, Z. Guan, W. Zhao, W. Lu, and J. Cui are with the State Key Laboratory of Integrated Services Networks, School of Computer Science and Technology, Xidian University, Xi'an, China. E-mail: \{yym@, zwang\_01@stu., zyguan@, ywzhao@mail., wglu@stu., cuijt@\}xidian.edu.cn}%
\thanks{X. Huang is the Key Laboratory of Intelligent Perception and Image Understanding of the Ministry of Education, the School of Artificial Intelligence, Xidian University, Xi'an, China. E-mail: xinyanh@stu.xidian.edu.cn}%
\thanks{X. He is with the State Key Laboratory of CAD\&CG, Zhejiang University, Hangzhou, China. E-mail: xiaofeihe@cad.zju.edu.cn}%
}

\markboth{Journal of \LaTeX\ Class Files, February~2025}%
{Yang \MakeLowercase{\textit{et al.}}: Aligning Multiple Knowledge Graphs in a Single Pass}


\maketitle

\begin{abstract}
Entity alignment (EA) is to identify equivalent entities across different knowledge graphs (KGs), which can help fuse these KGs into a more comprehensive one. Previous EA methods mainly focus on aligning a pair of KGs, and to the best of our knowledge, no existing EA method considers aligning multiple (more than two) KGs. To fill this research gap, in this work, we study a novel problem of aligning multiple KGs and propose an effective framework named MultiEA to solve the problem. First, we embed the entities of all the candidate KGs into a common feature space by a shared KG encoder. Then, we explore three alignment strategies to minimize the distances among pre-aligned entities. In particular, we propose an innovative inference enhancement technique to improve the alignment performance by incorporating high-order similarities. Finally, to verify the effectiveness of MultiEA, we construct two new real-world benchmark datasets and conduct extensive experiments on them. The results show that our MultiEA can effectively and efficiently align multiple KGs in a single pass. We release the source codes of MultiEA at: \url{https://github.com/kepsail/MultiEA}.
\end{abstract}

\begin{IEEEkeywords}
Knowledge Graphs, Entity Alignment, Graph Neural Networks
\end{IEEEkeywords}

\section{Introduction}
\IEEEPARstart{K}{nowledge} graphs (KGs) are a special kind of graph that can store a wealth of structural facts (i.e. knowledge) about the real world. Each fact is usually structured as a triple $(h,r,t)$, representing that head (subject) entity $h$ and tail (object) entity $t$ hold relation $r$ between them. In recent years, KGs have successfully supported many web applications such as search engines~\cite{zhao2021brain}, question-answer systems~\cite{embedkgqa,mrl-cqa,wang2017knowledge,k-eqa}, recommender systems~\cite{kgcn,kgat,wang2017knowledge}, knowledge graph reasoning~\cite{tpami-kg-reason,dura}, etc.

In practice, different KGs are constructed based on diverse data sources and different extraction approaches, and a single KG can usually cover only a specific aspect of structural facts. For example, an English KG usually contains more facts about the English-speaking society, while a Chinese KG contains more facts about the Chinese-speaking society. Considering that more KGs together can provide more comprehensive structural facts from various aspects, researchers have proposed many methods (Cf. surveys~\cite{openea,zhao2020experimental,zeng2023matching}) to fuse a pair of KGs into a unified one. The general process is to first identify equivalent entities between the candidate KGs, and then let them serve as the ``bridge entities'' to link the candidate KGs into a unified one.

Technically, existing mainstream methods typically solve this problem by embedding entities into a common latent representation space and minimizing the distance between pre-aligned entity pairs. This process is also referred to as entity alignment (EA). Depending on how to obtain the entity embeddings, they can be classified into two main categories: (1) Trans-based EA methods~\cite{jewp,iptranse,mtranse,jape,bootea,multike} learn entity embeddings by letting each triple satisfy some specific geometric properties in the embedding space. Most of these methods adopt TransE~\cite{transe} as the translation module to preserve the property: $h + r \approx t$. (2) GNN-based EA methods~\cite{gcn-align,avr-gcn,hman,hgcn-je-jr,gmnn,mugnn,nmn,ssp,rrea,psr,alinet,mraea,kecg,naea} adopt graph neural network (GNN) models such as graph convolutional network (GCN)~\cite{gcn} to learn entity embeddings by iteratively aggregating the embeddings of neighbor entities.

\begin{figure}
\centering
\includegraphics[width=0.95\columnwidth]{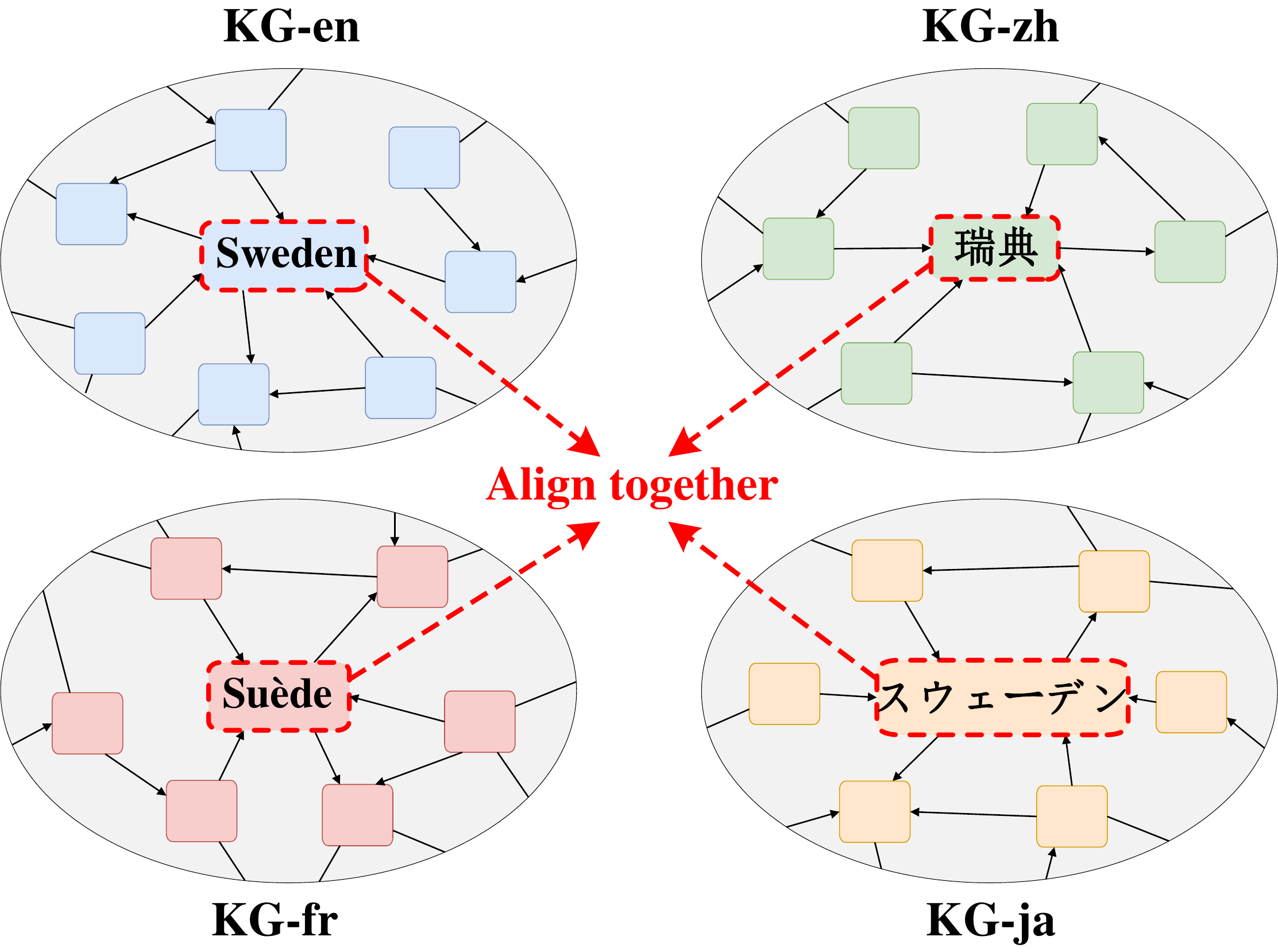}
\caption{An example of aligning four KGs.}
\label{fig:problem}
\end{figure}

However, all the existing EA methods are specifically designed for aligning only a pair of candidate KGs, which can cover limited knowledge for downstream applications. To the best of our knowledge, \textbf{none of the existing EA methods consider aligning multiple (more than two) KGs}. For example, Fig.~\ref{fig:problem} shows a case where four candidate KGs need to be aligned, which triggers a more challenging research problem. 

There are three obstacles that prevent traditional pair-wise KG alignment methods from being applied to the multiple KG alignment problem studied in this work.
First, existing EA datasets are particularly constructed for pair-wise KG alignment. At present, there is no benchmark dataset that can support the study of aligning multiple KGs.
Second, if we trivially adapt existing methods to this task, they need to be separately executed multiple times since they can only utilize the local (pair-wise) alignment information in a single pass, which is inefficient.
Last and most importantly, they cannot capture some useful global (beyond pair-wise) alignment information, affecting the final alignment performance. For the example in Fig.~\ref{fig:problem}, when they align KG-en with KG-fr, and align KG-en with KG-zh independently, the entities of the three KGs will be projected into four separate embedding spaces. Consequently, it is easy to yield inconsistent results like $e_1^{(en)} \equiv e_1^{(fr)}$, $e_1^{(en)} \equiv e_1^{(zh)}$, and $e_1^{(fr)} \equiv e_2^{(zh)}$ (wrong), violating the transitivity of the alignment relationship.

To fill the research gap in this field, in this work, we innovatively study the problem of aligning multiple KGs in a single pass. First of all, to facilitate the study of this problem, we construct two new benchmark datasets that contain multiple KGs to be aligned. Then, we formulate a novel research problem of aligning multiple KGs concurrently. Finally, we propose an effective framework named MultiEA to solve the problem. As illustrated in Fig.~\ref{fig:problem}, our MultiEA can concurrently align the four candidate KGs in a single pass, which is more practical. In addition, it embeds the entities of all the candidate KGs into a unified space and thus can capture useful global alignment information, which helps achieve better alignment performance. Recall the example above, if MultiEA pulls $e_1^{(en)}$ and $e_1^{(fr)}$ closer and pulls $e_1^{(en)}$ and $e_1^{(zh)}$ closer. Then, $e_1^{(fr)}$ and $e_1^{(zh)}$ will definitely be pulled closer as well, leading to the consistent result: $e_1^{(en)} \equiv e_1^{(fr)}$, $e_1^{(en)} \equiv e_1^{(zh)}$, and $e_1^{(fr)} \equiv e_1^{(zh)}$ (right). 

The main contributions of this work are summarized as follows.
\begin{itemize}
\item To the best of our knowledge, in the EA research community, we are the first to study the problem of concurrently aligning multiple (more than two) KGs. We are also the first to construct the benchmark datasets and define the evaluation metric for this problem.
\item We propose an effective framework named MultiEA to solve the problem. Three alignment strategies are explored to capture the global alignment information. An innovative inference enhancement technique is proposed to significantly boost the alignment performance.
\item We conduct extensive experiments on the two constructed benchmark datasets. The results demonstrate that MultiEA can effectively and efficiently align multiple KGs in a single pass. We release our source codes to facilitate further research on this problem.
\end{itemize}

\section{Related Work}
In this section, we comprehensively review existing methods that are related to ours.

\vspace{2mm} \textbf{Trans-based EA Methods.} Early EA methods exploit KG embedding methods~\cite{tpami-trans}, e.g., TransE~\cite{transe}, to project a pair of candidate KGs into a common low-dimensional Euclidean space by letting the embeddings of head entity $h$, relation $r$, and tail entity $h$ satisfy the triangular relationship: $h + r \approx t$. Given a set of pre-aligned entity pairs (i.e., seed alignment labels), they minimize the distance between these equivalent entity pairs in the embedding space. Thus, more potential equivalent entity pairs can be discovered after model optimization. Representative Trans-based EA methods include JEwP~\cite{jewp}, JAPE~\cite{jape}, MTransE~\cite{mtranse}, IPTransE~\cite{iptranse}, BootEA~\cite{bootea}, TransEdge~\cite{transedge}, MultiKE~\cite{multike}, etc. 

\vspace{2mm} \textbf{GNN-based EA Methods.} Motivated by the extraordinary success of graph neural networks (GNNs) in extracting the structural features of graphs, GCN-Align~\cite{gcn-align} first introduces GCN~\cite{gcn} into EA, which has shown superior performance than Trans-based EA Methods, e.g., JEwP~\cite{jewp}, MTransE~\cite{mtranse} and JAPE~\cite{jape}. Further, AVR-GCN~\cite{avr-gcn} deftly combines the graph convolutional operator in GNNs and the translation operator in Trans-based methods to extract better entity embeddings. Afterward, a series of following GNN-based EA methods, e.g., ~\cite{hman,hgcn-je-jr,gmnn,mugnn,nmn,ssp,alinet,mraea,naea,kecg,rrea,psr,largeea,tam2020entity,sun2022revisiting,peea} are proposed. Recently, RREA~\cite{rrea} insightfully summarizes KG embedding-based EA methods and GNN-based EA methods into a unified framework consisting of two modules: shape-builder and alignment. The former constraints KGs into a specific distribution in the embedding space, and the latter minimizes the distance between the embeddings of pre-aligned entity pairs.

\vspace{2mm} \textbf{EA Enhancement.} In recent years, several works further enhanced the EA performance by exploiting attribute information and entity name information. HMAN~\cite{hman}, BERT-INT~\cite{bert-int}, and ACK-MMEA~\cite{ack-mmea} can leverage the rich attributes of entities. Besides, PSR~\cite{psr}, GMNN~\cite{gmnn}, NMN~\cite{nmn}, HGCN-JE~\cite{hgcn-je-jr}, RDGCN~\cite{rdgcn}, LightEA~\cite{lightea} and SelfKG~\cite{selfkg} can use the word embeddings of the entity names to effectively initialize entity embeddings, greatly improving the performance of EA.

\vspace{2mm} \textbf{Alignment Methods in Other Fields.}
In the social network analysis field, there are several methods that have made some efforts to align multiple social networks. COSNET~\cite{cosnet} leverages multiple networks to enhance the alignment performance of two networks. It requires exponential time complexity to build a matching graph, which is impractical for many scenarios. ULink~\cite{ulink} learns a latent user space based on user attributes. It cannot exploit structural information and thus it is not a strict social network alignment approach. MC$^2$~\cite{mc2} combines both structural information and attribute information to infer a common base by matrix factorization. It requires all social networks to have attributes, and its time computational complexity is square to the number of uses. MASTER~\cite{master} embeds multiple social networks in a common latent space through collaborative matrix factorization, which also requires square computational complexity. In the data integration field, a recent work~\cite{multiem} is proposed to align multiple tables based on Sentence-BERT~\cite{reimers2019sentence}, table-wise hierarchical merging, and density-based pruning. These methods are specially designed methods for other fields and they generally require sophisticated optimization.

In summary, different from existing methods, in this work, we study the problem of aligning multiple (more than two) KGs based on the structural information. Accordingly, we develop a GNN-based neural network framework named MultiEA to solve the problem in an end-to-end optimization manner.

\section{Preliminaries}
\label{sec:prelim}
In this section, we first give the formal definition of knowledge graphs. Then, we formulate the novel problem of aligning multiple KGs.

\textbf{Knowledge Graph (KG).}
A knowledge graph is defined as $\mathcal{G} = (\mathcal{E}, \mathcal{R}, \mathcal{T})$, where $\mathcal{E}$ is the set of entities (nodes) and $\mathcal{R}$ is the set of relations (edges). $\mathcal{T} \subseteq \mathcal{E} \times \mathcal{R} \times \mathcal{E}$ is the set of triples, and each triple $t \in \mathcal{T}$ is represented as $<e_i, r_k, e_j>$, which means that head entity $e_i \in \mathcal{E}$ and tail entity $e_j \in \mathcal{E}$ hold relation $r_k \in \mathcal{R}$ between them. We use ${\mathbf h}_{i} \in {\mathbb R}^{d}$ to denote the representation of entity $e_i$ in one model layer, where $d$ is the dimensionality of embeddings. We use ${\mathbf g}_{k} \in {\mathbb R}^{d}$ to denote the embedding of relation $r_k$, which is a randomly initialized learnable parameter vector.

\textbf{Multiple KG Alignment.}
The input is a set of KGs $\{\mathcal{G}^{(1)}, \mathcal{G}^{(2)},  \mathcal{G}^{(m)}, ..., \mathcal{G}^{(M)}\}$, where $M$ is the number of the input KGs, and $M>2$. The $m$-th KG is denoted by $\mathcal{G}^{(m)} = (\mathcal{E}^{(m)}, \mathcal{R}^{(m)}, \mathcal{T}^{(m)})$.
The problem is to identify equivalent entities that refer to the same real-world thing across all the $M$ KGs, based on a set of seed alignment labels (i.e., pre-aligned entities), denoted as $\mathcal{S} = \{ l_1, l_2, ..., l_n, ..., l_N \}$, where $N$ is the number of labels. The $n$-th label is denoted as $l_n = (e_n^{(1)}, e_n^{(2)}, ..., e_n^{(m)}, ..., e_n^{(M)})$, where entity $e_n^{(m)} \in \mathcal{E}^{(m)}$, and all the $M$ entities associated with $l_n$ are equivalent.

As shown in Fig.~\ref{fig:problem}, KG-en, KG-zh, KG-fr, and KG-ja are four candidate KGs that need to be aligned, where the four entities marked with red dashed lines refer to the same object, i.e., ``Sweden''. We treat the four equivalent entities as one of the seed alignment labels, to help our EA algorithm find more potential equivalent entities. Finally, we can link (merge) the four KGs based on these detected equivalent entities across the four KGs.

\section{Framework}
In this section, we describe our proposed MultiEA framework in detail. Firstly, we design a GNN-like KG encoder to embed the entities of all the candidate KGs into a common vector space. Then, we propose three alignment strategies to measure the distances among pre-aligned entities, which guide the training of EA models. Finally, we develop an innovative inference enhancement technique to boost the alignment performance by incorporating high-order similarities.

\begin{figure*}[ht]
\centering
\subfigure[Mean Strategy]{\includegraphics[width=0.4\columnwidth]{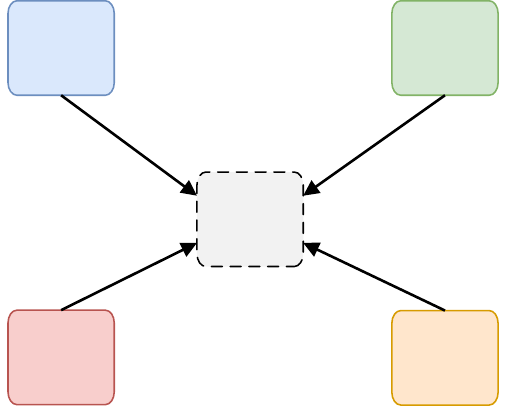}\label{fig:align-mean}}
\hspace{1cm}
\subfigure[Anchor Strategy]{\includegraphics[width=0.4\columnwidth]{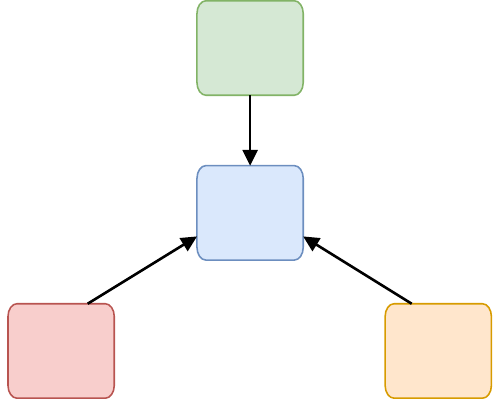}\label{fig:align-anchor}}
\hspace{1cm}
\subfigure[Each Other Strategy]{\includegraphics[width=0.45\columnwidth]{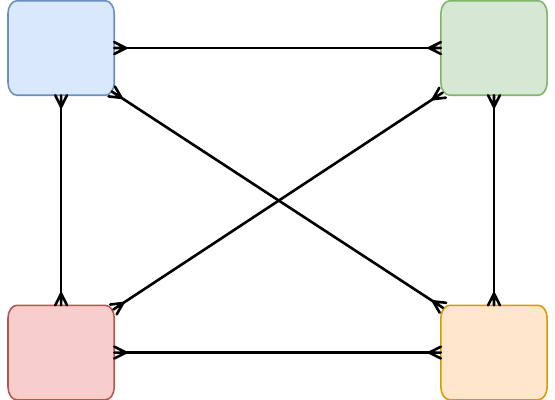}\label{fig:align-pair}}
\caption{
(a) Moving toward the mean (the central grey square with dashed line);
(b) Moving toward the anchor (the central blue square);
(c) Moving toward each other.
}
\label{fig:align}
\end{figure*}

\subsection{KG Encoding}
\label{sec:encoding}
Given a set of KGs: $\{\mathcal{G}^{(1)}, \mathcal{G}^{(2)}, ..., \mathcal{G}^{(m)}, ..., \mathcal{G}^{(M)}\}$, we use a shared KG-oriented GNN encoder to compute their entity embeddings. In the following, we take the $m$-th KG, i.e., $\mathcal{G}^{(m)} = (\mathcal{E}^{(m)}, \mathcal{R}^{(m)}, \mathcal{T}^{(m)})$ as an example to elaborate on the details of our encoder. The superscript $m$ is omitted for the sake of notation simplicity.

Firstly, we augment the original KG data to add a virtual self-relation $r_{self}$ into the relation set $\mathcal{R}$, formally described as follows:
\begin{equation}
\label{eq:aug-relation}
\mathcal{R} = \mathcal{R} \cup \{ r_{self} \}.
\end{equation}
Then, for each entity, we add a virtual triple to describe that each entity $e_i$ has the self-relation $r_{self}$ between itself, which is added to the triple set as follows:
\begin{equation}
\label{eq:aug-triple}
\mathcal{T} = \mathcal{T} \cup \{ <e_i, r_{self}, e_i> | e_i \in \mathcal{E}\}.
\end{equation}
To facilitate the aggregation of the GNN encoder, for entity $e_i$, we define the set of its neighbor relation-entity tuples as follows:
\begin{equation}
\label{eq:get-neighbor}
\mathcal{N}_i = \{ <r_{k}, e_j> | <e_i, r_k, e_j> \in \mathcal{T}, or <e_j, r_k, e_i> \in \mathcal{T} \}.
\end{equation}
Although the relations in KGs are usually directed, following most previous works, we treat them as bi-directed (or undirected). This is reasonable because the inverse of each relation is usually intuitive as well. For instance, for the triple $<$\textit{Obama}, \textit{wife}, \textit{Michelle}$>$, we can also rewrite it as: $<$\textit{Michelle}, \textit{husband}, \textit{Obama}$>$. Note that due to the addition of self-connection, entity $e_i$ will have a special neighbor relation-entity tuple: $<r_{self}, e_i> \in \mathcal{N}_i$.

In a layer of our KG encoder, the output representation of entity $e_i$ is computed as follows:
\begin{equation}
\label{eq:kg-aggregate}
{\mathbf h}_{i}' = \sigma \Big( \sum_{<r_k, e_j> \in \mathcal{N}_i} { {\alpha}_{i,k,j} \cdot {\mathbf W}_{k} \cdot {\mathbf h}_{j} } \Big),
\end{equation}
where ${\mathbf h}_{j}$ is the neighbor entity $e_j$'s feature vector output by the previous layer, ${\mathbf W}_{k}$ is a relation-specific projection matrix, ${\alpha}_{i,k,j}$ is the attention coefficient for aggregation, and $\sigma$ is the non-linear activation function.

There are often thousands of relations in common KGs, and thus the relation-specific projection matrices ${\mathbf W}_{k}, \forall r_k \in \mathcal{R}$ may introduce too many trainable parameters, risking overfitting. Therefore, following previous studies~\cite{rrea,psr}, we leverage the relation embedding vector ${\mathbf g}_{k}$ to define its related projection matrix, as follows:
\begin{equation}
\label{eq:rela-proj}
{\mathbf W}_{k}= {\mathbf I} - 2 \cdot {\mathbf g}_{k} \cdot {\mathbf g}_{k}^{T}.
\end{equation}
In this way, we no longer need to introduce additional parameters. In particular, the normalization constraint $\left \| {\mathbf g}_{k} \right \|_2 = 1$ is imposed, so that the orthogonality of ${\mathbf W}_{k}$ is naturally guaranteed since we can easily derive: \begin{equation}
{\mathbf W}_{k}^{T} \cdot {\mathbf W}_{k} = ({\mathbf I} - 2 \cdot {\mathbf g}_{k} \cdot {\mathbf g}_{k}^{T})^{T} ({\mathbf I} - 2 \cdot {\mathbf g}_{k} \cdot {\mathbf g}_{k}^{T}) = {\mathbf I}.
\end{equation}
The orthogonality of ${\mathbf W}_{k}$ is proven to be very beneficial for EA task~\cite{rrea,psr}.

The attention coefficient is computed as follows. Given a neighbor tuple $<r_k, e_j> \in \mathcal{N}_i$ of entity $e_i$, we first compute the proximity among the head entity $e_i$, the relation $r_k$, and the tail entity $e_i$:
\begin{equation}
\label{eq:attn-proximity}
{\beta}_{i,k,j} = \sigma \big( {\mathbf a}_h^T \cdot {\mathbf h}_{i} + {\mathbf a}_r^T \cdot {\mathbf g}_{k} + {\mathbf a}_t^T \cdot {\mathbf W}_{k} \cdot {\mathbf h}_{j} \big),
\end{equation}
where ${\mathbf a}_h$, ${\mathbf a}_r$, and ${\mathbf a}_t$ are learnable parameter vectors for head entities, relations, and tail entities, respectively. Then, we compute the attention coefficient ${\alpha}_{i,k,j}$ by normalizing ${\beta}_{i,k,j}$ across all the elements of $\mathcal{N}_i$:
\begin{equation}
\label{eq:attn-normalize}
{\alpha}_{i,k,j} = \frac{\exp \big( {\beta}_{i,k,j} \big)}{\sum_{<r_x, e_y> \in \mathcal{N}_i} \exp \big( {\beta}_{i,x,y} \big)}.
\end{equation}
Note that by adding the self-relation $r_{self}$, each entity $e_i$ itself acts as one of its own neighbors (Cf. Eqs. (\ref{eq:aug-relation}, \ref{eq:aug-triple})). In addition, we use the attention mechanism to aggregate all the neighbors (Cf. Eqs. (\ref{eq:kg-aggregate}, \ref{eq:attn-normalize})). Thus, as proven by~\cite{ie-hgcn}, our encoder is able to automatically learn the importance of arbitrary hops of neighborhood information. This is more flexible than~\cite{rrea,psr}, which directly concatenates all hops of neighborhood information.

Recall that the encoder is shared among all the KGs, and thus we can obtain the embeddings of the entities from every KG.

\subsection{Training for Alignment}
\label{sec:align}
In order to pull equivalent entities together in a unified embedding space, we adopt the following margin ranking loss as the training loss of our MultiEA model:
\begin{equation}
\label{eq:loss}
L=\sum_{l \in \mathcal{S}} \sum_{l' \in \mathcal{S}'} \max(d(l) - d(l') + \lambda, 0),
\end{equation}
where $\mathcal{S}$ is the set of positive examples, i.e., ground-truth alignment labels, $\mathcal{S}'$ is the set of negative examples, i.e., randomly generated false alignment labels, $d(\cdot)$ is a distance function, and  $\lambda > 0$ is a margin hyperparameter for separating positive and negative examples.

The negative examples in $\mathcal{S}'$ are generated as follows. For each positive example $l = (e^{(1)}, e^{(2)}, ..., e^{(m)}, ..., e^{(M)}) \in \mathcal{S}$, we fix one entity and replace the other entities by randomly sampling from their corresponding KGs, formally described as follows:
\begin{equation}
\label{eq:negative}
\eta \times
\left\{ 
\begin{array}{l}
(e^{(1)}, \tilde{e}^{(2)}, ..., \tilde{e}^{(m)}, ..., \tilde{e}^{(M)}), \\
(\tilde{e}^{(1)}, e^{(2)}, ..., \tilde{e}^{(m)}, ..., \tilde{e}^{(M)}), \\
..., \\
(\tilde{e}^{(1)}, \tilde{e}^{(2)}, ..., e^{(m)}, ..., \tilde{e}^{(M)}), \\
..., \\
(\tilde{e}^{(1)}, \tilde{e}^{(2)}, ..., \tilde{e}^{(m)}, ..., e^{(M)}) \\
\end{array} 
\right\},
\end{equation}
where $\tilde{e}^{(m)}$ is randomly sampled entity from $\mathcal{E}^{(m)}$ to replace the original entity $e^{(m)}$. Here, $\eta$ is an integer, indicating that for each positive example, we generate $\eta$ groups of negative examples in this way. Thus, each positive example corresponds to $\eta \times M$ negative examples.

In this work, we propose three strategies for minimizing the distance among equivalent entities. The ideas are intuitively shown in Fig.~\ref{fig:align}. For a positive example $l = (e^{(1)}, e^{(2)}, ..., e^{(m)}, ..., e^{(M)}) \in \mathcal{S}$, we firstly obtain its associated entity embeddings from the KG encoder and denote them as: $(\mathbf{h}^{(1)}, \mathbf{h}^{(2)}, ..., \mathbf{h}^{(m)}, ..., \mathbf{h}^{(M)})$. Then, the distance $d(l)$ is minimized by three strategies as follows.

\vspace{2mm} \textbf{(1) Moving toward Mean.}
This strategy lets all the equivalent entities approach their mean, as illustrated in Fig.~\ref{fig:align-mean}. We first compute the mean as follows:
\begin{equation}
\label{eq:mean}
\mathbf{u} = \frac{1}{M} \cdot \sum_{m=1}^{M} {\mathbf{h}^{(m)}}.
\end{equation}
Then, we sum the Euclidean distances between each entity embedding and the mean:
\begin{equation}
\label{eq:mean-dist}
d(l) = \sum_{m=1}^{M} {\|\mathbf{h}^{(m)} - \mathbf{u}\|_2}.
\end{equation}

\vspace{2mm} \textbf{(2) Moving toward Anchor.}
In practice, the candidate KGs are usually unbalanced due to various realistic factors. For instance, on our constructed multi-lingual KG dataset DBP-4, the English KG contains many more triples than the other three KGs (Cf. Table~\ref{tab:data}). Therefore, as shown in Fig.~\ref{fig:align-anchor}, we designate entities from KG DBP-4: en as anchor entities and let the equivalent entities from the other KGs approach these anchor entities. This idea is formally described as follows:
\begin{equation}
\label{eq:anchor-dist}
d(l) = \sum_{m=1}^{M} {\|\mathbf{h}^{(m)} - \mathbf{h}^{(a)}\|_2},
\end{equation}
where $\mathbf{h}^{(a)}$ is embedding of the anchor entity\footnote{Note that when $\mathbf{h}^{(m)}$ is designated as anchor entity, then $\|\mathbf{h}^{(m)} - \mathbf{h}^{(a)}\|_2 = 0$.}.

\vspace{2mm} \textbf{(3) Moving toward Each Other.}
This strategy lets equivalent entities approach each other. As illustrated in Fig.~\ref{fig:align-pair}, we minimize the distances between all possible pairs of equivalent entities, formally described as follows:
\begin{equation}
\label{eq:each-dist}
d(l) = \sum_{m_1 \neq m_2} {\|\mathbf{h}^{(m_1)} - \mathbf{h}^{(m_2)}\|_2},
\end{equation}
where $m_1$ and $m_2$ are in the set $\{ 1, 2, ..., m, ..., M \}$.

\begin{algorithm}[ht]
\caption{The training process of MultiEA.}
\label{alg:multiea}
\textbf{Input}: $M$ KGs $\{\mathcal{G}^{(1)}, \mathcal{G}^{(2)}, ..., \mathcal{G}^{(m)}, ..., \mathcal{G}^{(M)}\}$, and labels $\mathcal{S}$.\\
\begin{algorithmic}[1]
\STATE Randomly initialize model parameters;
\STATE Select one distance metric function from Eqs.~(\ref{eq:mean}, \ref{eq:mean-dist}), Eq.~(\ref{eq:anchor-dist}), or Eq.~(\ref{eq:each-dist});
\STATE Augment KG data by Eqs.~(\ref{eq:aug-relation}, \ref{eq:aug-triple});
\STATE Get neighbor relation-entity tuples by Eq.~(\ref{eq:get-neighbor});
\WHILE{not converge}
    \STATE Compute projection matrices by Eq.~(\ref{eq:rela-proj});
    \STATE Compute attention coefficients by Eqs.~(\ref{eq:attn-proximity}, \ref{eq:attn-normalize});
    \STATE Perform multiple layers of aggregation by Eq.~(\ref{eq:kg-aggregate});
    \STATE Randomly sample negative examples by Eq.~(\ref{eq:negative});
    \STATE Compute loss by Eq.~(\ref{eq:loss});
    \STATE Update model parameters by gradient descent;
\ENDWHILE
\end{algorithmic}
\end{algorithm}

\subsection{Training Time Complexity}
\label{subsec:complexity}
Let $|\mathcal{E}|$, $|\mathcal{R}|$, and $|\mathcal{T}|$ denote the total numbers of the entities, the relations, and the triples of all the $M$ candidate KGs, respectively. Let $d$ denote the dimensionalities of entity embeddings and relation embeddings. Let $|\mathcal{S}|$ denote the number of alignment labels. Recall that each label corresponds to $\eta \times M$ negative examples.

In the KG encoding phase (Cf. Section~\ref{sec:encoding}), the time complexity involves three main computational operations. First, the aggregation operation described by Eq.~(\ref{eq:kg-aggregate}) has time complexity ${\mathcal O}( |\mathcal{E}| \cdot d^2 )$. Second, the computation of relation-specific projection matrices described by Eq.~(\ref{eq:rela-proj}) has time complexity ${\mathcal O}( |\mathcal{R}| \cdot d^2 )$. Finally, as described by Eqs.~(\ref{eq:attn-proximity}, \ref{eq:attn-normalize}), the computation of attention coefficients has time complexity ${\mathcal O}( |\mathcal{T}| \cdot (d + d + d^2) )$. Considering that in practice, the embedding dimensionality $d$ is a relatively small value, the time complexity of KG encoding is ${\mathcal O}( |\mathcal{E}| + |\mathcal{R}| + |\mathcal{T}| )$. 

In the alignment phase (Cf. Section~\ref{sec:align}), as described by Eq.~(\ref{eq:mean-dist}) and Eq.~(\ref{eq:anchor-dist}), both the mean strategy and the anchor strategy require time complexity ${\mathcal O}(M)$. As described by Eq.~(\ref{eq:each-dist}), the each other strategy requires time complexity ${\mathcal O}(M^2)$. As described by Eq.~(\ref{eq:loss}), taking the number of labels into consideration. The mean strategy and the anchor strategy have time complexity of ${\mathcal O}(|\mathcal{S}| \cdot \eta \cdot M^2)$, and each other strategy has time complexity ${\mathcal O}(|\mathcal{S}| \cdot \eta \cdot M^3)$.

In practice, $|\mathcal{S}|$, $\eta$, and $M$ are much smaller than $|\mathcal{E}|$, $|\mathcal{R}|$, and $|\mathcal{T}|$. Therefore, the overall training time complexity of MultiEA is equal to ${\mathcal O}( |\mathcal{E}| + |\mathcal{R}| + |\mathcal{T}| )$.

\subsection{Inference Enhancement}
\label{sec:infer}
After training the model, we can obtain the embeddings of the entities from all the KGs. Based on these embeddings, we can compute the similarity between any two entities that are from different KGs. For example, given entity $e_{i}^{(m_1)} \in \mathcal{E}^{(m_1)}$, and entity $e_{j}^{(m_2)} \in \mathcal{E}^{(m_2)}$, we compute the similarity between them as follows\footnote{Since entity embeddings are normalized, i.e., $\left \| {\mathbf h_{i}} \right \|_2 = 1, \forall i$, the computed similarities lie in the range $[0,1]$. Thus, we can compose high-order similarities by matrix product operation later.}:
\begin{equation}
\label{eq:infer-sim}
{\mathbf S}_{i,j}^{m_1 - m_2} = 1 - \frac{\|\mathbf{h}_{i}^{(m_1)} - \mathbf{h}_{j}^{(m_2)}\|_2}{2}.
\end{equation}

Thus, we can obtain a matrix ${\mathbf S}^{m_1 - m_2}$ to describe the entity similarities between $\mathcal{G}^{(m_1)}$ and $\mathcal{G}^{(m_2)}$. However, Eq.~(\ref{eq:infer-sim}) only captures the first-order similarity, which may not be sufficient since some higher-order similarity information is ignored. Therefore, in this work, we propose to further enhance the similarity matrix by incorporating higher-order similarities, which can be composed by matrix product operation. For example, if there are three KGs, $\mathcal{G}^{(m_1)}$, $\mathcal{G}^{(m_2)}$, and $\mathcal{G}^{(m_3)}$, we can enhance the similarity matrix ${\mathbf S}^{m_1 - m_2}$ as follows:
\begin{equation}
\label{eq:infer-3}
{\widetilde{\mathbf S}}^{m_1 - m_2} = \gamma_1 \cdot {\mathbf S}^{m_1 - m_2} + \gamma_2 \cdot {\mathbf S}^{m_1 - m_3} \cdot {\mathbf S}^{m_3 - m_2},
\end{equation}
where $\gamma_1$ and $\gamma_2$ are hyperparameters to balance the two terms. They are real numbers in the range $[0,1]$ and satisfy $\gamma_1 + \gamma_2 = 1$. Similarly, if there are four KGs, we can enhance ${\mathbf S}^{m_1 - m_2}$ as follows:
\begin{equation}
\label{eq:infer-4}
\begin{split}
{\widetilde{\mathbf S}}^{m_1 - m_2} &= \gamma_1 \cdot {\mathbf S}^{m_1 - m_2} + \gamma_2 \cdot {\mathbf S}^{m_1 - m_3} \cdot {\mathbf S}^{m_3 - m_2} \\
&+ \gamma_3 \cdot {\mathbf S}^{m_1 - m_4} \cdot {\mathbf S}^{m_4 - m_2},
\end{split}
\end{equation}
where $\gamma_1$, $\gamma_2$, and $\gamma_3$ are hyperparameters in the range $[0,1]$ and satisfy $\gamma_1 + \gamma_2 + \gamma_3 = 1$.

Here, we only incorporate the two-order similarities by the matrix product operation. In practice, higher-order similarities can be similarly incorporated through more matrix product operations. Finally, we can identify equivalent entities according to the enhanced similarity matrix. In this way, the model can utilize more comprehensive information, helping improve the inference performance.

\section{Experiment}
\label{sec:experiment}
In this section, we construct the benchmark dataset and define the evaluation metric for the new task. Then, we conduct extensive experiments to verify the effectiveness of our MultiEA.

\subsection{Benchmark Datasets}
\label{sec:data-detail-brief}
As far as we know, there is no existing EA benchmark dataset that can support the task of aligning multiple KGs. To this end, we construct two novel benchmark datasets based on previously widely used real-world datasets~\cite{jape, bootea}. One dataset DBP-4 contains four candidate KGs, and another dataset DWY-3 contains three candidate KGs to be aligned. The key statistics of the constructed datasets are listed in Table~\ref{tab:data}.

\begin{table}[ht]
\tabcolsep=0.3cm
 \renewcommand{\arraystretch}{1.3}
\begin{center}
\caption{Dataset statistics.}
\label{tab:data}
\begin{tabular}{ccccc@{\extracolsep{0.8cm}}c}
\toprule
Datasets & KGs & $|\mathcal{E}|$ & $|\mathcal{R}|$ & $|\mathcal{T}|$ & $|\mathcal{S}|$ \\
\midrule
\multirow{4}[0]{*}{DBP-4} & en & 8901 & 1034 & 77483 & \multirow{4}[0]{*}{2539} \\
 & fr & 3545 & 774 & 15843 & \\
 & ja & 4326 & 519 & 32427 & \\
 & zh & 3893 & 619 & 21497 & \\
\midrule
\multirow{3}[0]{*}{DWY-3} & dbp & 23784 & 246 & 94985  & \multirow{3}[0]{*}{20729} \\
 & wiki & 22839 & 153 & 92019 & \\
 & yago & 22063 & 30 & 77457 & \\
\bottomrule
\end{tabular}
\end{center}
\end{table}

\textbullet\
\textbf{DBP-4.}
This dataset contains four KGs from different language sources, including English (en) KG, Chinese (zh) KG, Japanese (ja) KG, and French (fr) KG. It is constructed based on the DBP15K dataset, which was originally released by~\cite{jape}. The original alignment labels describe the correspondence from English KG to the other three KGs, which can be described as: en-fr, en-zh, and en-ja. We treat the English KG as an intermediary and complement the alignment information among all four KGs, which can be described as: en-fr-zh-ja. Specifically, we filter entities from different KGs that correspond to the same English entity. Intuitively, if we have: $e^{(en)} = e^{(fr)}$, $e^{(en)} = e^{(zh)}$, and $e^{(en)} = e^{(ja)}$. Then, we can infer: $e^{(en)} = e^{(fr)} = e^{(zh)}= e^{(ja)}$, which means that all these four entities from different KGs are equivalent.

\textbullet\
\textbf{DWY-3.}
This dataset contains three KGs, i.e., DBpedia (dbp), Wikidata (wiki), and YAGO (yago). It is constructed based on the DWY100K dataset, which was originally released by~\cite{bootea}. Its original alignment labels describe the correspondence from DBpedia to the other two KGs, which can be described as: dbp-wiki, and dbp-yago. Similar to the construction of DBP-4, we treat the DBpedia KG as an intermediary to complement the alignment information among all three KGs, which can be described as: dbp-wiki-yago.

It is worth noting that the condition under which we construct labels is very strict. As a result, the number of labels is very small compared to the number of entities and triples. To address the issue, we further induce a smaller knowledge graph. Specifically, inspired by the previous work~\cite{jape}, for each entity involved in labels, we select its popular neighbor entities whose degrees are larger than a threshold (we set 15 in this work). This target entity and its high-degree neighbor entries are added to a new entity set, and the involved relations are added to a new relation set. Then, we induce a new set of triples based on the selected entities and relations.

\subsection{Evaluation Metric}
Traditional pair-wise KG alignment methods generally use $Hits@K$ to evaluate the alignment performance. Specifically, given two KGs: $\mathcal{G}^{(l)}$ and $\mathcal{G}^{(r)}$, we first consider aligning $\mathcal{G}^{(r)}$ to $\mathcal{G}^{(l)}$, and define a temporary metric $l\_Hits@K$ to measure the proportion of the entities in $\mathcal{G}^{(l)}$ whose counterpart in $\mathcal{G}^{(r)}$ rank in top-K:
\begin{equation}
\label{eq:l-hits}
l\_Hits@K = \frac{\sum_{n=1}^{N} {\mathbb{I}(e_n^{(r)}\text{ranks in top-K})}}{N},
\end{equation}
where $\mathbb{I}$ is the indicator function that outputs 1 if the input fact is true and $0$ otherwise. Then, we consider aligning $\mathcal{G}^{(l)}$ to $\mathcal{G}^{(r)}$, and similarly define a temporary metric $r\_Hits@K$ in the opposite direction. The final metric $Hits@K$ is computed as the average of the two temporary metrics:
\begin{equation}
\label{eq:hits-k}
Hits@K = \frac{1}{2} \cdot (l\_Hits@K + r\_Hits@K).
\end{equation}

Unfortunately, $Hits@K$ does not work for our task of multiple KG alignment. Hence, in the spirit of $Hits@K$, we define a novel evaluation metric. Given $M$ KGs: $\{\mathcal{G}^{(1)}, \mathcal{G}^{(2)}, ..., \mathcal{G}^{(m)}, ..., \mathcal{G}^{(M)}\}$ where $M>2$, we treat each KG $\mathcal{G}^{(m)}$ as the target KG, and consider aligning the other $M-1$ KGs to it. First, we count the number of entities from $\mathcal{G}^{(m)}$ whose \textbf{all} $M-1$ counterparts from the other $M-1$ KGs rank in top-K, described as follows:
\begin{equation}
\begin{split}
C = \sum_{n=1}^{N} {\widetilde{\mathbb{I}}(e_n^{(1)}...,e_n^{(m-1)},e_n^{(m+1)}...,e_n^{(M)} \text{rank in top-K})},
\end{split}
\end{equation}
where $\widetilde{\mathbb{I}}(\cdot)$ is an indicator function that outputs $1$ if \textbf{all} the input entities rank in top-K, and $0$ otherwise. Then, a temporary metric $m\_Hits@K$ is defined to compute the proportion of:
\begin{equation}
m\_Hits@K = \frac{C}{N}.
\end{equation}
The final metric is computed as the average of all the $M$ temporary metrics: 
\begin{equation}
\label{eq:m-hits-k}
M\text{-}Hits@K = \frac{1}{M} \cdot \sum_{m=1}^{M} {m\_Hits@K}.
\end{equation}
Obviously, our newly defined metric $M\text{-}Hits@K$ is more challenging than the traditional metric $Hits@K$ since the condition for counting a correct entity is much more strict. This can better reflect the model performance of aligning multiple KGs.

\begin{table*}[ht]
\tabcolsep=0.36cm
\renewcommand{\arraystretch}{1.3}
\begin{center}
\caption{Comparison of multiple KG alignment accuracy. The best results are \textbf{bolded}, and the second best results are \underline{underlined}. The notation $^*$ denotes that we re-implement the baseline methods to make them compatible with the task.}
\label{tab:multi-kg-acc}
\begin{tabular}{ccccccc}
\toprule
\multirow{2}[0]{*}{Methods} & \multicolumn{3}{c}{DBP-4} & \multicolumn{3}{c}{DWY-3} \\
& $M\text{-}Hits@1$ & $M\text{-}Hits@10$ & $M\text{-}Hits@20$ & $M\text{-}Hits@1$ & $M\text{-}Hits@10$ & $M\text{-}Hits@20$ \\
\midrule
MTransE$^*$ & 3.40\% & 19.72\% & 32.64\% & 25.83\% & 55.94\% & 65.31\% \\
IPTransE$^*$ & 4.11\% & 23.36\% & 39.55\% & 28.74\% & 60.02\% & 68.84\% \\
GCN-Align$^*$ & 4.48\% & 27.62\% & 44.94\% & 32.46\% & 63.18\% & 73.91\% \\
KECG$^*$ & 5.62\% & 33.35\% & 49.61\% & 34.08\% & 67.51\% & 77.52\% \\
NAEA$^*$ & 6.14\% & 34.13\% & 48.07\% & 35.17\% & 69.88\% & 79.43\% \\
PSR$^*$ & 7.84\% & 38.17\% & 50.97\% & 37.55\% & 71.85\% & 80.13\% \\
\midrule
\textbf{MultiEA}+\textit{mean}-\textit{infer} & 5.95\% & 40.48\% & 51.02\% & 36.68\% & 69.33\% & 79.65\% \\
\textbf{MultiEA}+\textit{anchor}-\textit{infer} & 6.32\% & 40.33\% & 52.08\% & 37.54\% & 70.21\% & 79.19\% \\
\textbf{MultiEA}+\textit{each}-\textit{infer} & 6.17\% & 41.70\% & 52.00\% & 36.76\% & 71.52\% & 79.18\% \\
\textbf{MultiEA}+\textit{mean}+\textit{infer} & 10.03\% & 48.18\% & 59.89\% & 40.54\% & 73.90\% & 81.24\% \\
\textbf{MultiEA}+\textit{anchor}+\textit{infer} & \textbf{10.58\%} & \underline{48.80\%} & \underline{59.95\%} & \textbf{42.33\%} & \underline{74.13\%} & \underline{83.54\%} \\
\textbf{MultiEA}+\textit{each}+\textit{infer} & \underline{10.25\%} & \textbf{50.38\%} & \textbf{61.75\%} & \underline{41.19\%} & \textbf{75.44\%} & \textbf{85.42\%} \\
\bottomrule
\end{tabular}
\end{center}
\end{table*}

\subsection{Baselines and Variants}
Regarding baselines, we select six representative supervised EA baselines that only leverage the structural information of candidate KGs, including two Trans-based EA methods (1), (2), and four GNN-based EA methods (3), (4), (5), (6). They are listed as follows:

\begin{enumerate}
\item \textbf{MTransE}~\cite{mtranse} is a Trans-based EA method that uses TransE~\cite{transe} to encode entities of each KG in a separated embedding space;
\item \textbf{IPTransE}~\cite{iptranse} further uses PTransE~\cite{ptranse} to encode KGs and propose an iterative alignment strategy;
\item \textbf{GCN-Align}~\cite{gcn-align} is the first GNN-based EA method that uses GCN~\cite{gcn} to encode two KGs;
\item \textbf{NAEA}~\cite{naea} further uses GAT~\cite{gat} to encode the structural information of KGs. Therefore, it learns entity embeddings by aggregating neighbor entities with different importance;
\item \textbf{KECG}~\cite{kecg} also uses GAT~\cite{gat} to encode KGs, and additionally restricts the projection matrix to be a diagonal matrix, reducing the number of parameters and computations;
\item \textbf{PSR}~\cite{psr} encodes KGs based on orthogonal projection matrix and attention aggregation like ours. Differently, (1) it dose not add self-connections, and (2) it directly concatenates all hops of neighborhood information.
\end{enumerate}

Recall that in Section~\ref{sec:align}, we introduce three different alignment strategies. Here, we use +\textit{mean}, +\textit{anchor}, and +\textit{each} to denote the three strategies, respectively. In Section~\ref{sec:infer}, we propose the inference enhancement module. Here, we use +\textit{infer} or -\textit{infer} to indicate whether the module is equipped. By considering all the permutations, we set the following six variants for our MultiEA:

\begin{enumerate}
\item \textbf{MultiEA}+\textit{mean}-\textit{infer};
\item \textbf{MultiEA}+\textit{anchor}-\textit{infer};
\item \textbf{MultiEA}+\textit{each}-\textit{infer};
\item \textbf{MultiEA}+\textit{mean}+\textit{infer};
\item \textbf{MultiEA}+\textit{anchor}+\textit{infer};
\item \textbf{MultiEA}+\textit{each}+\textit{infer}.
\end{enumerate}

For anchor-based variants, we treat DBP-4: en and DWY-3: dbp as the anchor KGs on the two datasets, respectively.

\subsection{Implementation Details}
\label{sec:detail}
For the constructed benchmark datasets of DBP-4 and DWY-3, we follow the convention to randomly select 30\% elements of the seed alignment set $\mathcal{S}$ to form the training set, and the rest are treated as the test set. We implement our MultiEA based on a PyTorch-based open-source EA toolkit named EAkit~\cite{eakit}. For baselines, we use their implementations provided in the EAkit package. All the experiments are conducted on our constructed benchmark datasets, and all the methods use the same dataset partitioning.

For our MultiEA, we use the same hyperparameter settings for both datasets. This is a challenging configuration since it can better reflect the sensitivity of hyperparameters w.r.t. datasets. All the model parameters are randomly initialized by the Xavier uniform distribution~\cite{xavier}. We adopt the Adam optimizer, and the learning rate is set to 0.01. The embedding dimensionalities of both entities and relations are set to 256. The non-linear activation function $\sigma$ is set to the ELU function~\cite{elu_nonlinear}. The margin hyperparameter $\lambda$ is set to 1. The number of negative example groups, i.e., the hyperparameter $\eta$ is set to 10. We adopt early stopping for model training, and the number of patience steps is set to 10.

All the experiments are conducted on an NVIDIA TITAN RTX GPU with 24GB GPU memory and 128GB main memory.

\begin{figure*}[ht]
\centering
\includegraphics[width=\linewidth]{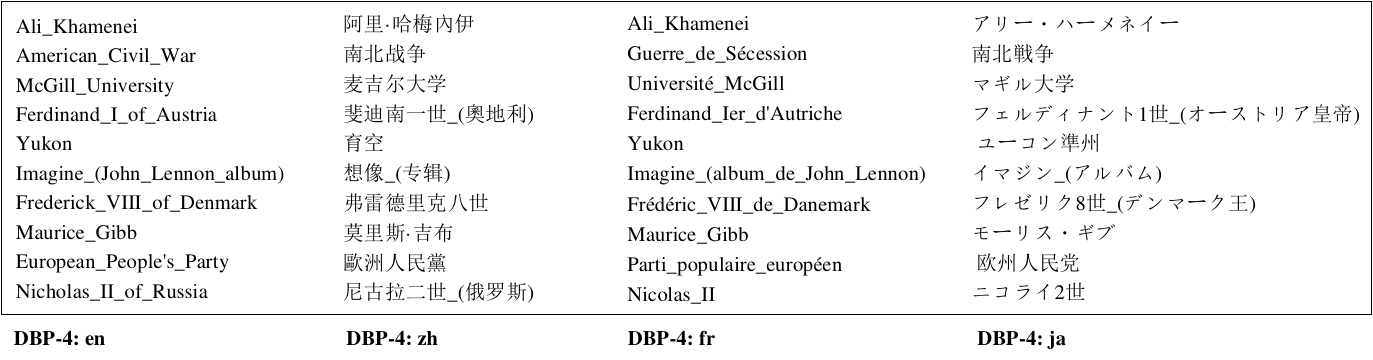}
\caption{The entity names of the ten groups of equivalent entities discovered by MultiEA on DBP-4.}
\label{fig:case-dbp}
\end{figure*}

\begin{figure*}[!ht]
\centering
\includegraphics[width=\linewidth]{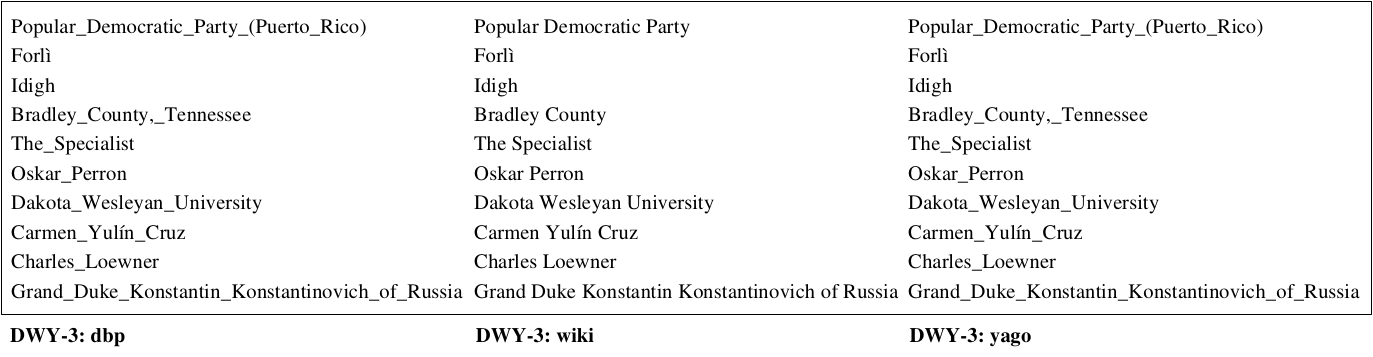}
\caption{The entity names of ten groups of equivalent entities discovered by MultiEA on DWY-3. Note that, for DWY-3: wiki, the original data is in the form of the URLs of the entities. We show the corresponding entries of the involved URLs.}
\label{fig:case-dwy}
\end{figure*}

\begin{figure*}[!ht]
\centering
\subfigure[DBP-4]{\includegraphics[width=0.85\columnwidth]{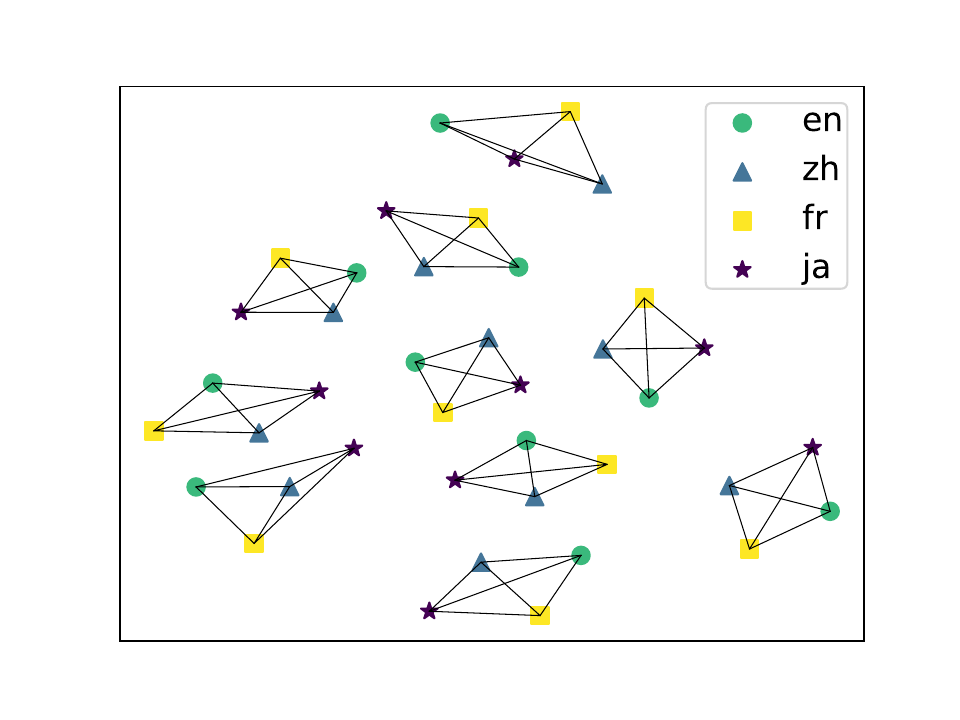}\label{fig:visual_dbp}}
\subfigure[DWY-3]{\includegraphics[width=0.85\columnwidth]{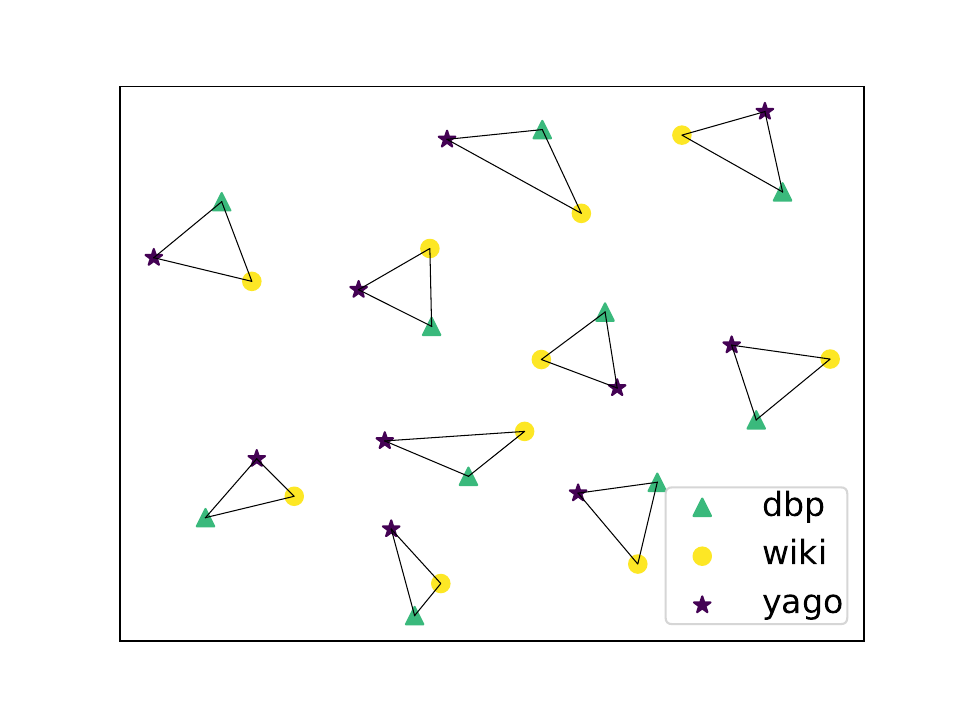}\label{fig:visual_dwy}}
\caption{Visualization of the embeddings of the ten groups of equivalent entities listed in Fig.~\ref{fig:case-dbp} and Fig.~\ref{fig:case-dwy}.}
\label{fig:visual}
\end{figure*}

\begin{table*}[ht]
\tabcolsep=0.4cm
\renewcommand{\arraystretch}{1.3}
\begin{center}
\caption{Comparison of accuracy and efficiency on DBP-4.}
\label{tab:multi-kg-acc-dbp-4}
\begin{tabular}{ccccccc}
\toprule
Methods & $M\text{-}Hits@1$ & $M\text{-}Hits@10$ & $M\text{-}Hits@20$ & Parameter & Memory & Time \\
\midrule
GCN-Align$^*$ & 4.48\% & 27.62\% & 44.94\% & 10.04M & 38.30MB & 382.77s \\
\textbf{MultiEA} (GCN-Align) & \textbf{7.32}\% & \textbf{43.75}\% & \textbf{54.14}\% & \textbf{5.35}M & \textbf{20.43}MB & \textbf{145.49}s \\
\midrule
KECG$^*$ & 5.62\% & 33.35\% & 49.61\% & 10.24M & 39.07MB & 478.30s \\
\textbf{MultiEA} (KECG) & \textbf{8.42}\% & \textbf{45.93}\% & \textbf{57.60}\% & \textbf{5.42}M & \textbf{20.68}MB & \textbf{185.66}s \\
\midrule
NAEA$^*$ & 6.14\% & 34.14\% & 48.07\% & 12.31M & 46.95MB & 522.60s \\
\textbf{MultiEA} (NAEA) & \textbf{8.95}\% & \textbf{46.15}\% & \textbf{57.82}\% & \textbf{6.43}M & \textbf{24.56}MB & \textbf{206.13}s \\
\midrule
PSR$^*$ & 7.84\% & 38.17\% & 50.97\% & 11.13M & 42.47MB & 1336.05s \\
\textbf{MultiEA} (PSR) & \textbf{9.98}\% & \textbf{49.13}\% & \textbf{61.30}\% & \textbf{6.05}M & \textbf{23.06}MB & \textbf{576.19}s \\
\bottomrule
\end{tabular}
\end{center}
\end{table*}

\begin{table*}[ht]
\tabcolsep=0.4cm
\renewcommand{\arraystretch}{1.3}
\begin{center}
\caption{Comparison of accuracy and efficiency on DWY-3.}
\label{tab:multi-kg-acc-dwy-3}
\begin{tabular}{ccccccc}
\toprule
Methods & $M\text{-}Hits@1$ & $M\text{-}Hits@10$ & $M\text{-}Hits@20$ & Parameter & Memory & Time \\
\midrule
GCN-Align$^*$ & 32.46\% & 63.18\% & 73.91\% & 23.80M & 90.80MB & 566.43s \\
\textbf{MultiEA} (GCN-Align) & \textbf{36.75}\% & \textbf{69.75}\% & \textbf{80.26}\% & \textbf{17.78}M & \textbf{67.83}MB & \textbf{315.44}s \\
\midrule
KECG$^*$ & 34.08\% & 67.51\% & 77.52\% & 24.04M & 91.66MB & 780.29s \\
\textbf{MultiEA} (KECG) & \textbf{38.52}\% & \textbf{71.85}\% & \textbf{82.97}\% & \textbf{17.78}M & \textbf{67.83}MB & \textbf{407.16}s \\
\midrule
NAEA$^*$ & 35.17\% & 69.88\% & 79.43\% & 25.63M & 93.96MB & 868.15s \\
\textbf{MultiEA} (NAEA) & \textbf{40.66}\% & \textbf{74.69}\% & \textbf{83.66}\% & \textbf{18.31}M & \textbf{69.84}MB & \textbf{480.61}s \\
\midrule
PSR$^*$ & 37.55\% & 71.85\% & 80.13\% & 23.85M & 90.98MB & 1974.62s \\
\textbf{MultiEA} (PSR) & \textbf{41.68}\% & \textbf{74.76}\% & \textbf{84.72}\% & \textbf{17.58}M & \textbf{67.07}MB & \textbf{804.37}s \\
\bottomrule
\end{tabular}
\end{center}
\end{table*}

\subsection{Multiple KG Alignment}
\label{subsec:experim-multi-kg}
We first quantitatively evaluate the effectiveness of our MultiEA in aligning multiple (more than two) KGs. Specifically, we apply all six baseline methods as well as all six variants of MultiEA to the two constructed benchmark datasets, and use the metric $M\text{-}Hits@K$ newly defined in Eq.~(\ref{eq:m-hits-k}) to compare their accuracy. The results are reported in Table~\ref{tab:multi-kg-acc}.

\vspace{2mm} \textbf{Comparison with Baselines.} It is worth noting that we cannot directly apply baseline EA methods to the multiple KG alignment task since they are specifically designed for the traditional pair-wise KG alignment task. To make them compatible, we split the concerned task into multiple pair-wise sub-tasks and separately align each pair of KGs. To reflect this, baselines are marked by $^*$ in the table. We can see that our MultiEA (especially the three variants with the inference enhancement module at the bottom of the table) can significantly outperform all the baselines on the two datasets. This may be due to that our MultiEA can effectively capture the global alignment information by projecting the entities of all the candidate KGs in a unified embedding space. Besides, GNN-based EA baselines show better performance than Trans-based EA baselines. This is consistent with the findings of many previous studies~\cite{gcn-align,kecg,naea,psr}.

\vspace{2mm} \textbf{Abalation Studies.} As we can see, our MultiEA+\textit{each}+\textit{infer} achieves the best overall results in most cases, showing its superior effectiveness in aligning multiple KGs. Overall, MultiEA+\textit{anchor}+\textit{infer} and MultiEA+\textit{mean}+\textit{infer} achieve the second-best and the third-best results, respectively. The situation is the same for the other three variants without the inference enhancement module. Therefore, we recommend the ``each other" strategy as the default strategy for most cases. If users prefer higher efficiency and there is an obvious anchor KG, we recommend the ``anchor" strategy, and otherwise, we recommend the ``mean" strategy. From another dimension, we can observe that all three variants with the inference enhancement module significantly outperform the other three variants without this module. This indicates that the inference enhancement module is significantly beneficial for this task.

\subsection{Case Study}
\label{sec:visual}
To intuitively show the effectiveness of our MultiEA, in Fig.~\ref{fig:case-dbp} and Fig.~\ref{fig:case-dwy}, we list ten groups of equivalent entities discovered by MultiEA (the best-performing variant MultiEA+\textit{each}+\textit{infer}) on the two datasets. As we can see, MultiEA can effectively discover equivalent entities across all the candidate KGs.

Further, we visualize the embeddings of these entities. Specifically, we utilize the t-SNE algorithm~\cite{tsne} to project them into the 2-dimensional Euclidean space. Fig.~\ref{fig:visual_dbp} and Fig.~\ref{fig:visual_dwy} show the results on DBP-4 and DWY-3, respectively, where different colors and different shapes mark the entities from different KGs, and the equivalent entities are connected by black lines. As we can see, on both datasets, the ten groups of equivalent entities are obviously gathered together, indicating the effectiveness of the alignment mechanism of MultiEA. On the other hand, a proper margin is maintained between different groups, avoiding the risk of model overfitting. The result on DWY-3 shows a relatively better distribution. This is consistent with the higher accuracy on DWY-3 as reported in Table~\ref{tab:multi-kg-acc}.

\subsection{Efficiency Study}
\label{sec:effic}
In this experiment, we demonstrate the efficiency advantage of our MultiEA in aligning multiple KGs. As shown in Table~\ref{tab:multi-kg-acc-dbp-4} and Table~\ref{tab:multi-kg-acc-dwy-3}, we first report the experimental results of our re-implemented GNN-based EA baselines as described in Section~\ref{subsec:experim-multi-kg}. That is, they align multiple candidate KGs by splitting this task into multiple separate pair-wise sub-tasks, as marked by $^*$ in the tables. Then, we apply the ``each other'' alignment strategy and the inference enhancement module of our MultiEA+\textit{each}+\textit{infer} to each baseline X, as denoted as MultiEA (X) in the tables. 

In addition to the $M\text{-}Hits@K$ scores, we report the number of model parameters, the memory, and the time required by each method during its training. As we can see, our MultiEA can help each baseline achieve significantly better performance. Moreover, MultiEA can help each baseline significantly reduce the requirements of parameters, space, and time resources. This demonstrates the superior efficiency of our MultiEA.

\begin{table*}[ht]
\tabcolsep=0.5cm
 \renewcommand{\arraystretch}{1.3}
\begin{center}
\caption{Pair-wise KG alignment $Hits@1$ score. MultiEA (X) denotes combining our training strategy with the X encoder.}
\label{tab:pair-kg-acc-hits1}
\begin{tabular}{cccccc}
\toprule
\multirow{2}[0]{*}{Methods} & \multicolumn{3}{c}{DBP-4} & \multicolumn{2}{c}{DWY-3} \\
 & en-zh & en-fr & en-ja & dbp-wiki & dbp-yago \\
\midrule
MTransE & 14.63\% & 15.91\% & 13.54\% & 35.16\% & 44.35\% \\
\textbf{MultiEA} (MTransE) & \textbf{15.29\%} & \textbf{16.05\%} & \textbf{15.68\%} & \textbf{36.05\%} & \textbf{46.17\%} \\
Relative Gain $\uparrow$ & +4.51\% & +0.88\% & +15.81\% & +2.53\% & +4.10\% \\
\midrule
IPTransE & 19.84\% & 20.95\% & 21.38\% & 43.51\% & 55.19\% \\
\textbf{MultiEA} (IPTransE) & \textbf{20.26\%} & \textbf{21.01\%} & \textbf{23.29\%} & \textbf{43.68\%} & \textbf{57.06\%} \\
Relative Gain $\uparrow$ & +0.11\% & +0.33\% & +8.93\% & +0.39\% & +3.38\% \\
\midrule
GCN-Align & 21.48\% & 24.01\% & 21.54\% & 44.75\% & 58.63\% \\
\textbf{MultiEA} (GCN-Align) & \textbf{21.57\%} & \textbf{25.71\%} & \textbf{22.19\%} & \textbf{45.48\%} & \textbf{60.08\%} \\
Relative Gain $\uparrow$ & +0.42\% & +7.08\% & +3.02\% & +1.63\% & +2.47\% \\
\midrule
KECG & 25.71\% & 27.44\% & 26.77\% & 47.15\% & 61.97\% \\
\textbf{MultiEA} (KECG) & \textbf{26.14\%} & \textbf{29.48\%} & \textbf{27.49\%} & \textbf{50.02\%} & \textbf{62.44\%} \\
Relative Gain $\uparrow$ & +1.67\% & +7.43\% & +2.69\% & +6.09\% & +0.76\% \\
\midrule
NAEA & 24.69\% & 26.92\% & 26.35\% & 46.51\% & 61.37\% \\
\textbf{MultiEA} (NAEA) & \textbf{25.75\%} & \textbf{28.16\%} & \textbf{26.95\%} & \textbf{49.18\%} & \textbf{61.52\%} \\
Relative Gain $\uparrow$ & +4.29 & +4.61 & +2.28 & +5.74 & +0.24 \\
\midrule
PSR & 25.59\% & 28.20\% & 27.87\% & 48.23\% & 60.52\% \\
\textbf{MultiEA} (PSR) & \textbf{26.27\%} & \textbf{30.18\%} & \textbf{28.03\%} & \textbf{51.36\%} & \textbf{62.14\%} \\
Relative Gain $\uparrow$ & +2.65\% & +7.02\% & +0.57\% & +6.49\% & +3.12\% \\
\bottomrule
\end{tabular}
\end{center}
\end{table*}

\begin{table*}[!ht]
\tabcolsep=0.5cm
 \renewcommand{\arraystretch}{1.3}
\begin{center}
\caption{Pair-wise KG alignment $Hits@10$ scores. MultiEA (X) denotes combining our training strategy with the X encoder.}
\label{tab:pair-kg-acc-hits10}
\begin{tabular}{cccccc}
\toprule
\multirow{2}[0]{*}{Methods} & \multicolumn{3}{c}{DBP-4} & \multicolumn{2}{c}{DWY-3} \\
 & en-zh & en-fr & en-ja & dbp-wiki & dbp-yago \\
\midrule
MTransE & 45.52\% & 50.13\% & 47.93\% & 60.54\% & 67.09\% \\
\textbf{MultiEA} (MTransE) & \textbf{45.93\%} & \textbf{52.26\%} & \textbf{50.19\%} & \textbf{60.96\%} & \textbf{67.36\%} \\
Relative Gain $\uparrow$ & +0.90\% & +4.25\%  & +4.72\% & +0.69\% & +0.40\% \\
\midrule
IPTransE & 56.23\% & 57.79\% & 60.66\% & 75.96\% & 79.32\% \\
\textbf{MultiEA} (IPTransE) & \textbf{58.52\%} & \textbf{59.35\%} & \textbf{62.43\%} & \textbf{77.16\%} & \textbf{82.17\%} \\
Relative Gain $\uparrow$ & +4.07\% & +2.70\% & +2.92\% & +1.58\% & +3.59\% \\
\midrule
GCN-Align & 62.03\% & 68.05\% & 63.96\% & 77.58\% & 86.00\% \\
\textbf{MultiEA} (GCN-Align) & \textbf{62.41\%} & \textbf{70.37\%} & \textbf{65.22\%} & \textbf{78.56\%} & \textbf{88.02\%} \\
Relative Gain $\uparrow$ & +0.61\% & +3.41\% & +1.97\% & +1.26\% & +2.35\% \\
\midrule
KECG & 67.19\% & 71.29\% & 67.51\% & 80.40\% & 88.65\% \\
\textbf{MultiEA} (KECG) & \textbf{68.03\%} & \textbf{73.10\%} & \textbf{69.38\%} & \textbf{80.73\%} & \textbf{89.78\%} \\
Relative Gain $\uparrow$ & +1.25\% & +2.54\% & +2.77\% & +0.41\% & +1.27\% \\
\midrule
NAEA & 66.41 & 71.43 & 67.83 & 80.67 & 88.76 \\
\textbf{MultiEA} (NAEA) & \textbf{68.18\%} & \textbf{74.17\%} & \textbf{70.95\%} & \textbf{81.35\%} & \textbf{89.37\%} \\
Relative Gain $\uparrow$ & +2.67\% & +3.84\% & +4.60\% & +0.84\% & +0.69\% \\
\midrule
PSR & 67.83\% & 73.04\% & 69.93\% & 82.41\% & 90.26\% \\
\textbf{MultiEA} (PSR) & \textbf{70.04\%} & \textbf{75.93\%} & \textbf{71.81\%} & \textbf{83.97\%} & \textbf{90.79\%} \\
Relative Gain $\uparrow$ & +3.25\% & +3.96\% & +2.69\% & +1.90\% & +0.59\% \\
\bottomrule
\end{tabular}
\end{center}
\end{table*}

\subsection{Pair-wise KG Alignment}
\label{sec:pair-align}
We also apply our MultiEA (the variant MultiEA+\textit{each}+\textit{infer}) to the traditional pair-wise KG alignment task. Firstly, based on the two constructed benchmark datasets, we induce several pair-wise KG alignment sub-datasets. For DBP-4, we induce three sub-datasets, i.e., DBP-4: en-fr, DBP-4: en-ja, and DBP-4: en-zh. For DWY-3, we induce two sub-datasets, i.e., DWY-3: dbp-wiki and DWY-3: dbp-yago. For baselines, we reproduce their results on our induced datasets. Then, we re-implement the encoder of MultiEA as that used in a baseline X, denoted as MultiEA (X). This can eliminate the encoder's influence on the alignment results, and thus investigate whether our proposed ``each other" strategy and inference enhancement module helps improve the performance of baselines on this task. By convention, we use the widely used metric $Hits@K$ described by Eq.~(\ref{eq:hits-k}) to evaluate the alignment accuracy.

Table~\ref{tab:pair-kg-acc-hits1} and Table~\ref{tab:pair-kg-acc-hits10} shows the final results. In addition to the original experimental results, we also compute the relative gain to show the improvement more intuitively. As we can see, our MultiEA can always help improve the baselines' performance on the traditional pair-wise KG alignment task. Besides, we have some other findings as follows. PSR, KECG, and NAEA perform better than GCN-Align, indicating the effectiveness of the attention mechanism used in their KG encoders. Moreover, MultiEA (PSR), MultiEA (KECG), and MultiEA (NAEA) outperform MultiEA (GCN-Align). This demonstrates that our MultiEA framework can effectively exploit the structural information captured by the graph attention mechanisms. Two Trans-based methods MTransE and IPTransE show worse performance than GNN-based methods, which is consistent with the conclusions of previous works~\cite{gcn-align,kecg,naea}. 

The experimental results reported here are slightly different from the results reported in previous studies. The main reason lies in that our constructed datasets are different from existing datasets. For example, the label ratio on our DBP-4 is $\frac{2539 \times 4}{8901+3545+4326+3893} \approx 0.4915$, while the label ratio of DBP15K used in previous studies is $(\frac{15000 \times 2}{19388+19572}+\frac{15000 \times 2}{19780+19814}+\frac{15000 \times 2}{19993+19661}) \times \frac{1}{3} \approx 0.7614$, which is much larger than ours.

\subsection{Hyperparameter Study}
\label{subsec:hyper}

\begin{figure*}[!ht]
\centering
\subfigure[$\gamma$ on DBP-4]{\includegraphics[width=0.5\columnwidth]{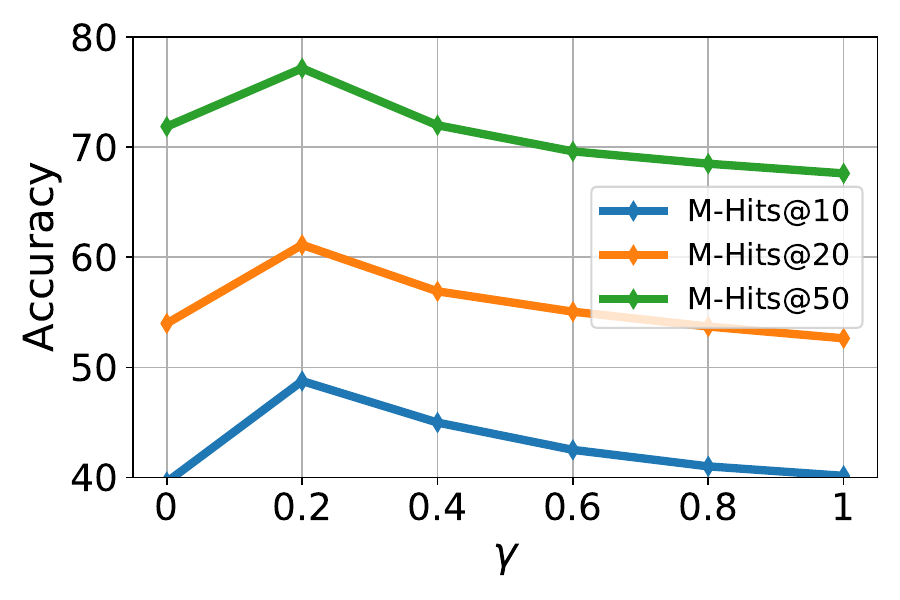}\label{fig:gamma_dbp}}
\subfigure[$\lambda$ on DBP-4]{\includegraphics[width=0.5\columnwidth]{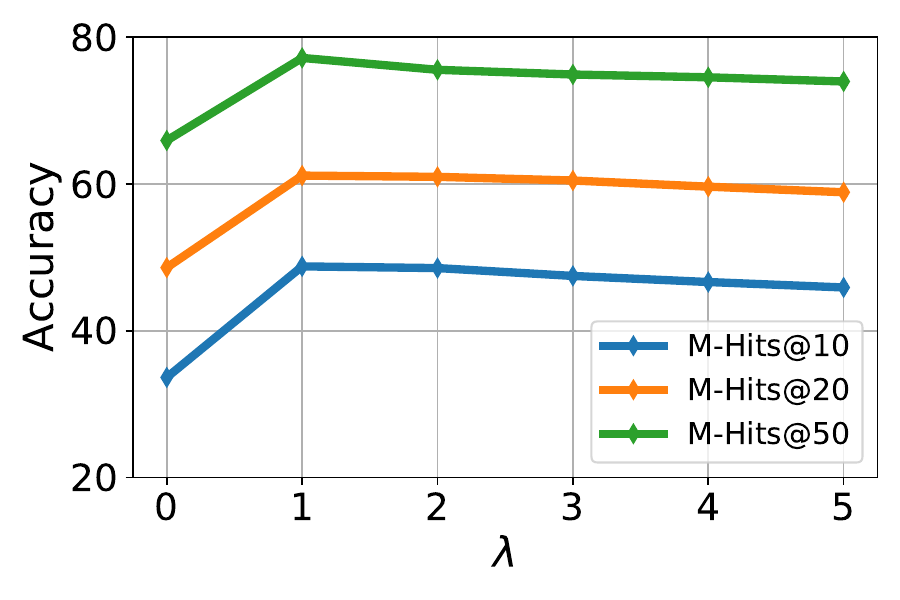}\label{fig:lambda_dbp}}
\subfigure[$\eta$ on DBP-4]{\includegraphics[width=0.5\columnwidth]{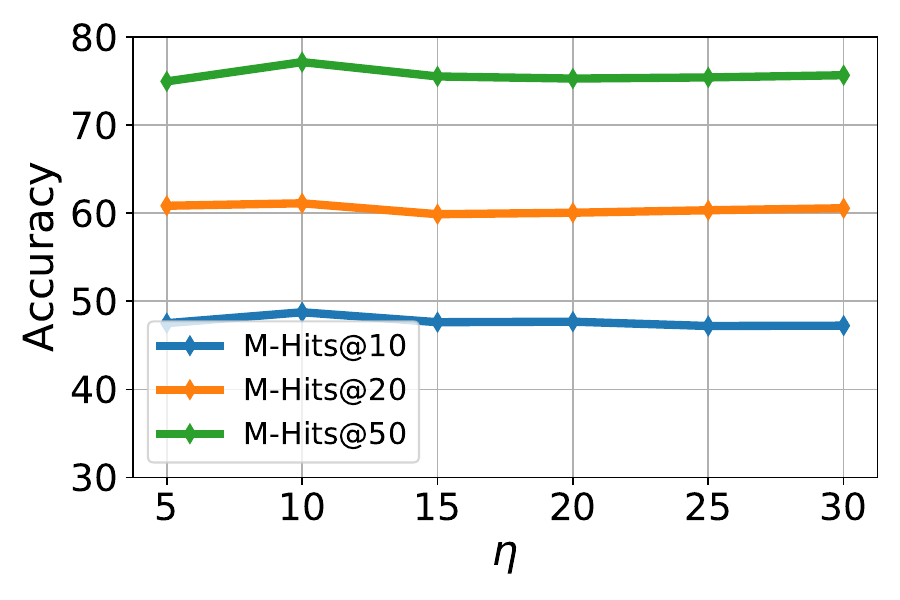}\label{fig:negative_dbp}}
\subfigure[ratio on DBP-4]{\includegraphics[width=0.5\columnwidth]{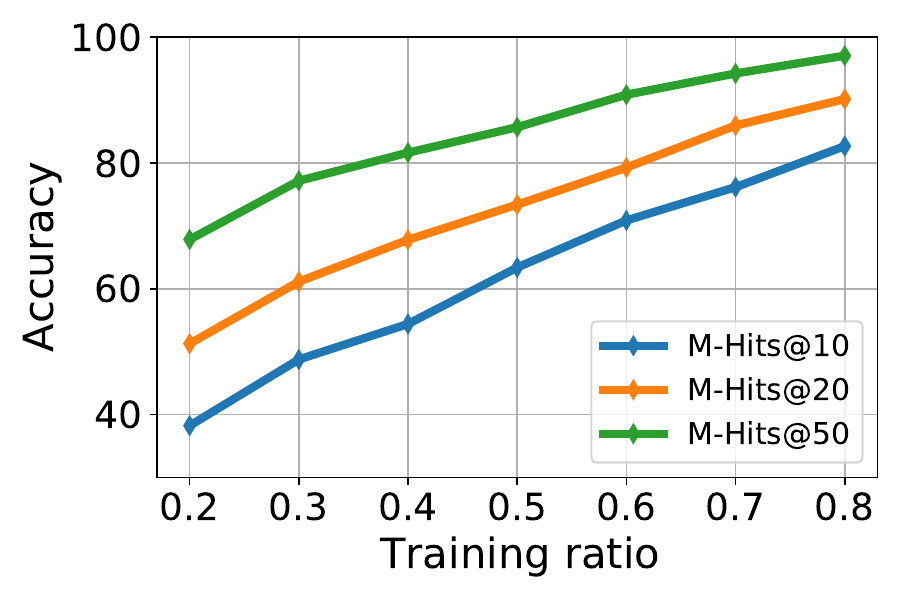}\label{fig:training_dbp}}\\
\subfigure[$\gamma$ on DWY-3]{\includegraphics[width=0.5\columnwidth]{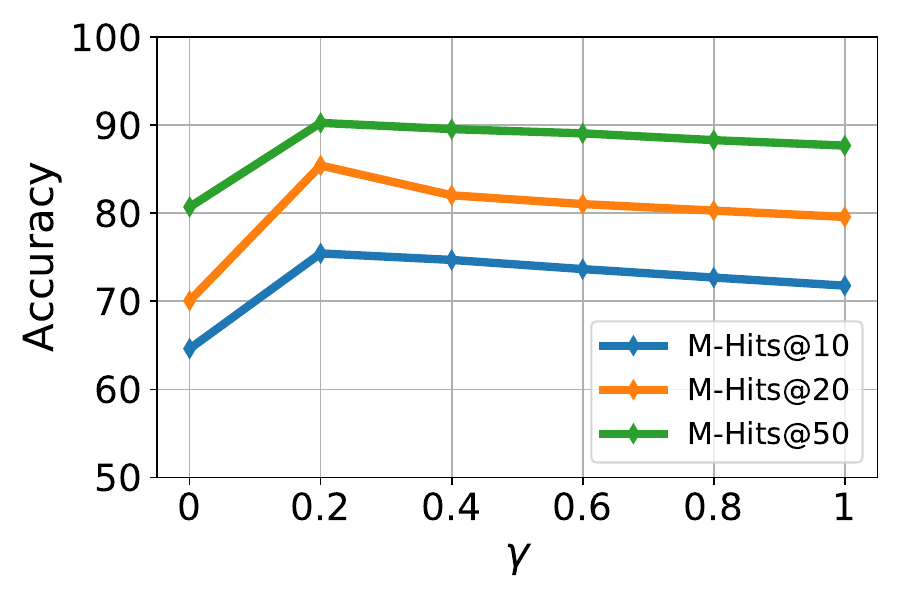}\label{fig:gamma_dwy}}
\subfigure[$\lambda$ on DWY-3]{\includegraphics[width=0.5\columnwidth]{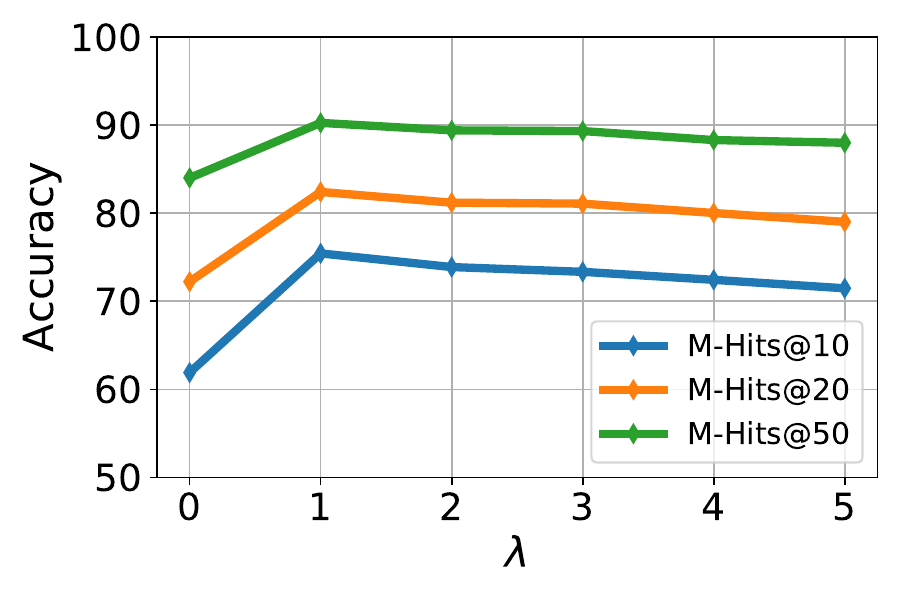}\label{fig:lambda_dwy}}
\subfigure[$\eta$ on DWY-3]{\includegraphics[width=0.5\columnwidth]{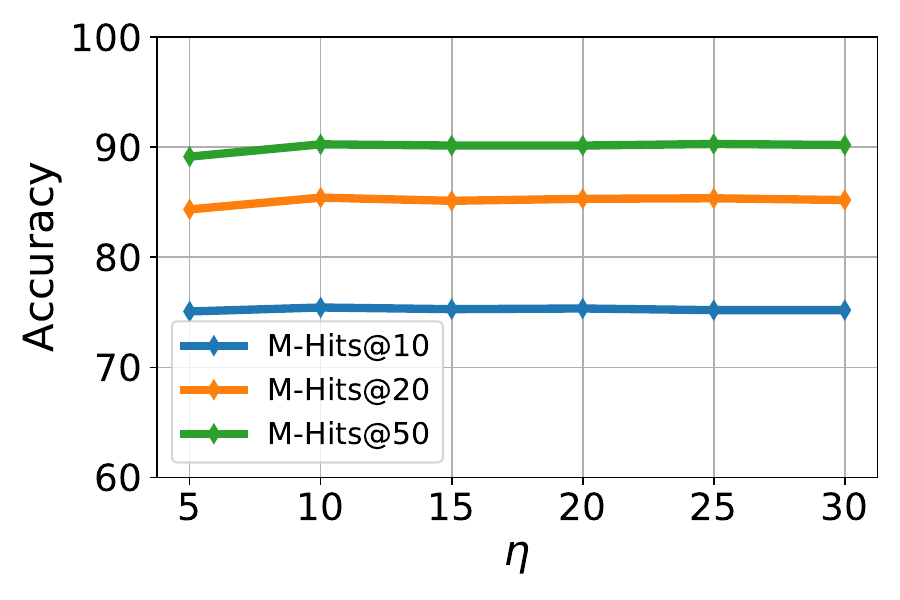}\label{fig:negative_dwy}}
\subfigure[ratio on DWY-3]{\includegraphics[width=0.5\columnwidth]{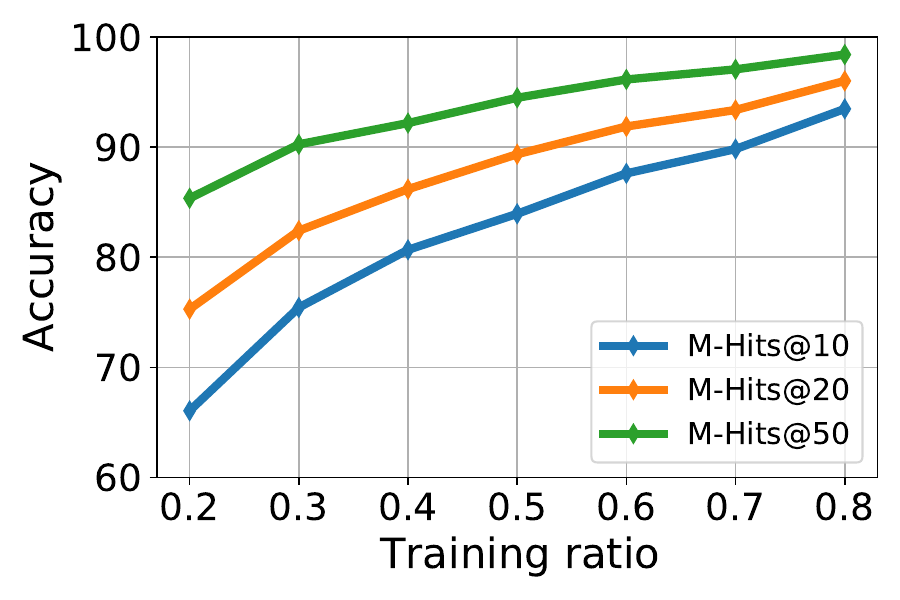}\label{fig:training_dwy}}
\caption{The sensitivity analysis of key hyperparameters.}
\label{fig:hyperparam}
\end{figure*}

In this subsection, we study the impact of MultiEA's four key hyperparameters on multiple KG alignment accuracy.

First, we study the balance hyperparameters introduced in the inference enhancement module, as described by Eqs.~(\ref{eq:infer-3}, \ref{eq:infer-4}). In practice, we only search the weight for the first term, and the weights for the other terms are set to the same value, thus compressing the hyperparameter search space. For one example of DBP-4 dataset, we can rewrite Eq.~(\ref{eq:infer-4}) as: 
\begin{equation}
\begin{split}
{\widetilde{\mathbf S}}^{en - zh} &= \gamma \cdot {\mathbf S}^{en - zh} + \frac{1 - \gamma}{2} \cdot {\mathbf S}^{en - fr} \cdot {\mathbf S}^{fr - zh} \\ 
&+ \frac{1 - \gamma}{2}  \cdot {\mathbf S}^{en - ja} \cdot {\mathbf S}^{ja - zh}.\nonumber 
\end{split}
\end{equation}
For another example of DWY-3 dataset, we can rewrite Eq.~(\ref{eq:infer-3}) as: 
\begin{equation}
{\widetilde{\mathbf S}}^{dbp - wiki} = \gamma \cdot {\mathbf S}^{dbp - wiki} + (1 - \gamma) \cdot {\mathbf S}^{dbp - yago} \cdot {\mathbf S}^{yago - wiki}.\nonumber
\end{equation} 
The sensitivity of $\gamma$ on the two datasets is shown in Fig.~\ref{fig:gamma_dbp} and Fig.~\ref{fig:gamma_dwy}, respectively. We can see that on both datasets, our MultiEA achieves the best performance when $\gamma = 0.2$. It means that the importance of the first-order similarity is 0.2 and the importance of the other higher-order similarities is 0.8. This indicates that incorporating high-order similarities is very beneficial for the multiple KG alignment task, demonstrating the effectiveness of our proposed inference enhancement module. This finding is also consistent with the observation in Section~\ref{subsec:experim-multi-kg}.

Second, we study the sensitivity of the margin hyperparameter $\lambda$ introduced in the loss function as described by Eq.~(\ref{eq:loss}). The results are shown in Fig.~\ref{fig:lambda_dbp} and Fig.~\ref{fig:lambda_dwy}. As we can see, the performance is very poor when $\lambda = 0$, because the model cannot separate the positive examples and the negative examples, causing the underfitting issue. There is a clear inflection point when $\lambda = 1$, and thus in the other experiments, we set it to 1 by default. After that, the performance gradually declines, probably because the model suffers the overfitting issue when the margin is too large.

Third, we investigate the sensitivity of $\eta$, which has been introduced to describe the groups of negative examples. As shown in Fig.~\ref{fig:negative_dbp} and Fig.~\ref{fig:negative_dwy}, our model is not sensitive to this hyperparameter. Therefore, in practice, we set $\eta$ to 10 to save computational resources.

Finally, we show the model performance w.r.t. the training ratio in Fig.~\ref{fig:training_dbp} and Fig.~\ref{fig:training_dwy}. As expected, the training ratio shows a clear positive effect on improving the model performance, which is consistent with the findings of many previous studies, e.g.,~\cite{jape,hgcn-je-jr,kecg,bootea,rdgcn,psr}.

\section{Conclusion}
In this work, we note that existing methods mainly focus on aligning only a pair of candidate KGs, ignoring the multiplicity of the candidate KGs to be aligned. To fill the research gap in this field, we formulate a novel problem of aligning multiple (more than two) KGs and propose the MultiEA framework to effectively solve this problem. To enhance the inference performance, we innovatively propose to incorporate higher-order similarities. Last but not least, we construct two new benchmark datasets and define the evaluation metric for the problem. The experimental results show that our MultiEA framework can effectively and efficiently align multiple KGs in a single pass. We will make our source codes and the constructed benchmark datasets publicly available to facilitate further research on this problem, which we believe will inspire more interesting works in the future.

\section*{Acknowledgments}
This work was supported in part by the National Natural Science Foundation of China under Grants 62133012, 61936006, 62425605, and 62303366, and in part by the Key Research and Development Program of Shaanxi under Grant 2024CY2-GJHX-15.

\bibliographystyle{IEEEtran}
\bibliography{multiea}

\begin{IEEEbiography}[{\includegraphics[width=0.9in, height=1.25in, clip, keepaspectratio]{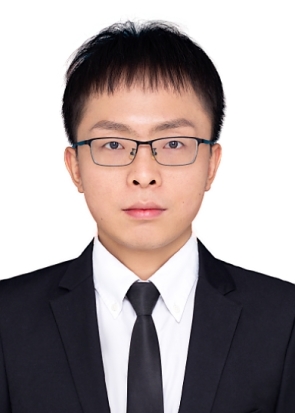}}]{Yaming Yang} received the B.S. and Ph.D. degrees in Computer Science and Technology from Xidian University, China, in 2015 and 2022, respectively. He is currently a lecturer with the School of Computer Science and Technology at Xidian University. His research interests include data mining and machine learning on graph data.
\end{IEEEbiography}

\begin{IEEEbiography}[{\includegraphics[width=0.9in, height=1.25in, clip, keepaspectratio]{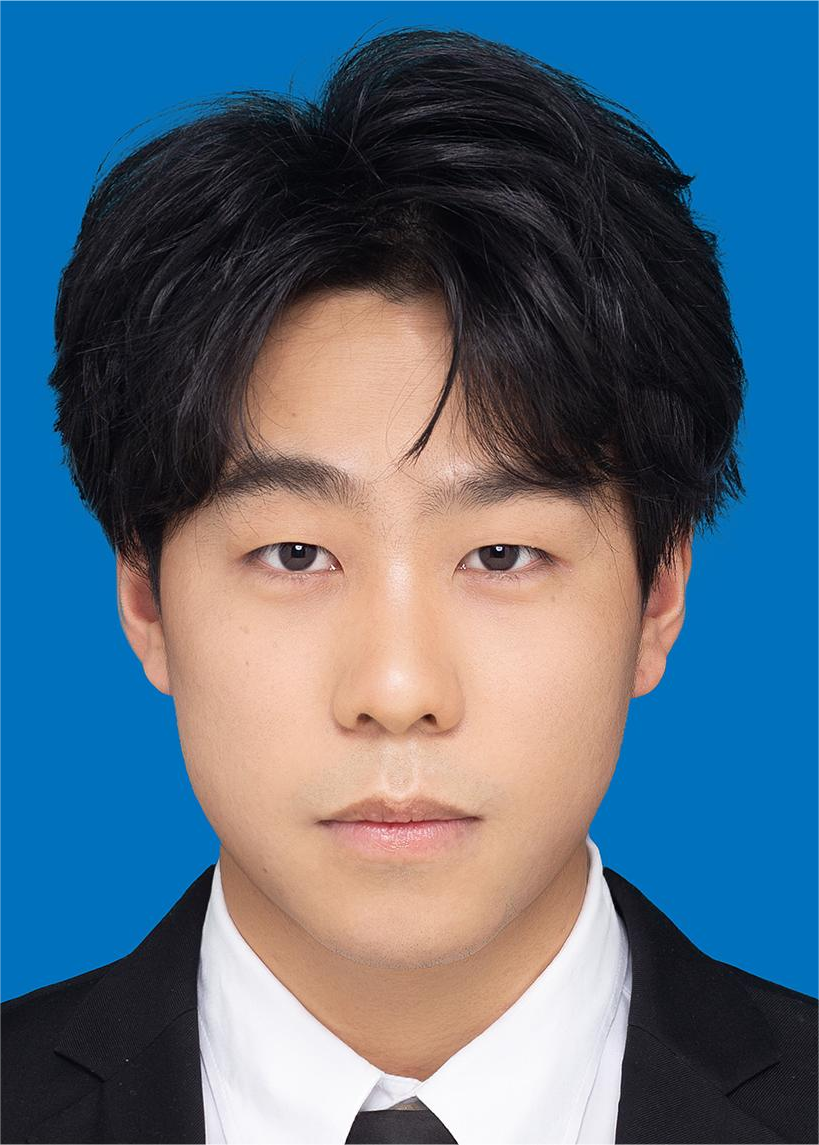}}]{Zhe Wang} received the B.S. degree in Management from Hefei University of Technology, China, in 2020. He is currently working towards a Ph.D. degree with the School of Computer Science and Technology, Xidian University, China. His research interests include data mining and machine learning on knowledge graph data.
\end{IEEEbiography}

\begin{IEEEbiography}[{\includegraphics[width=0.9in, height=1.25in, clip, keepaspectratio]{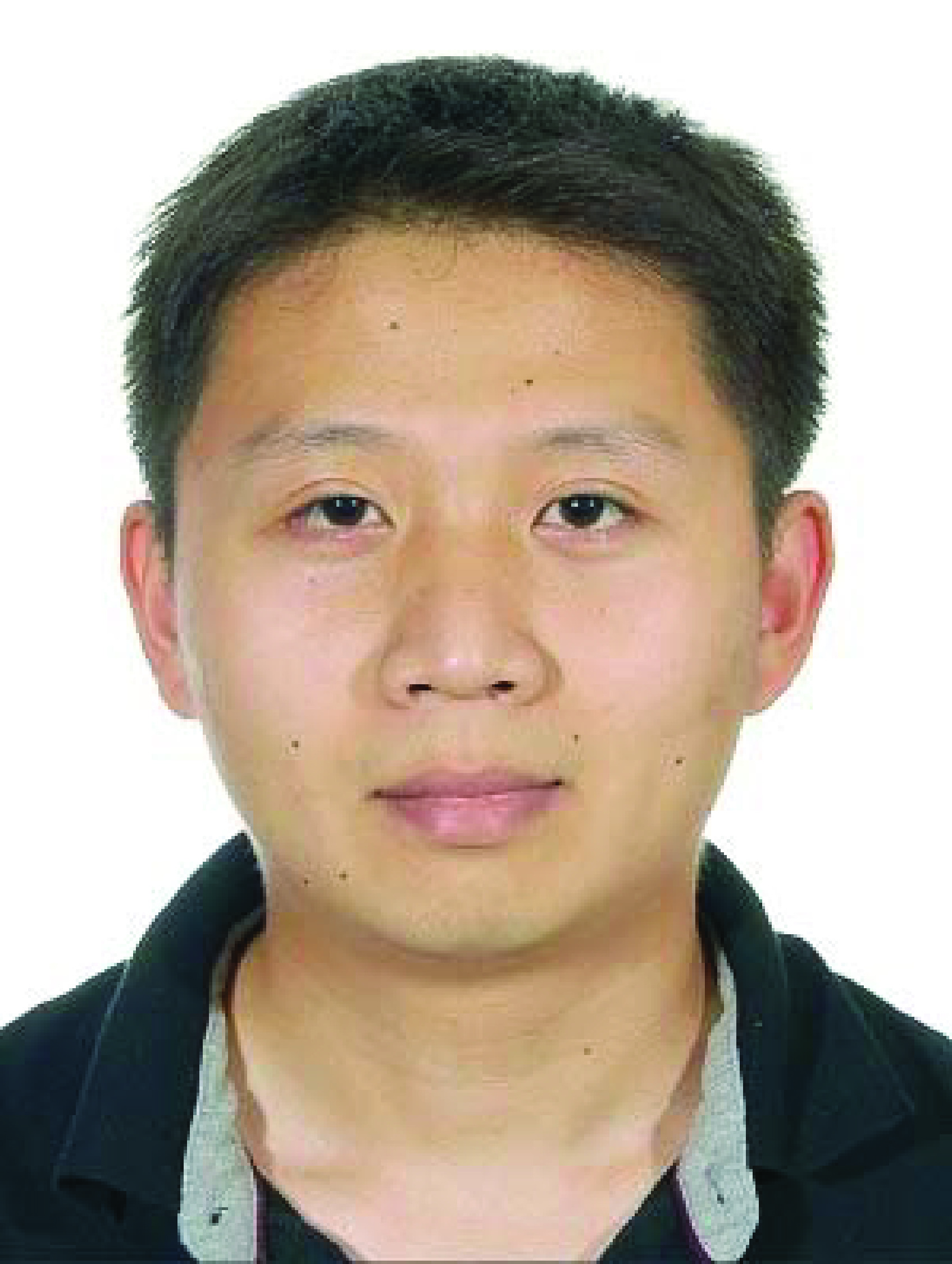}}]{Ziyu Guan} received the B.S. and Ph.D. degrees in Computer Science from Zhejiang University, Hangzhou China, in 2004 and 2010, respectively. He had worked as a research scientist in the University of California at Santa Barbara from 2010 to 2012, and as a professor in the School of Information and Technology of Northwest University, China from 2012 to 2018. He is currently a professor with the School of Computer Science and Technology, Xidian University. His research interests include attributed graph mining and search, machine learning, expertise modeling and retrieval, and recommender systems.
\end{IEEEbiography}

\begin{IEEEbiography}[{\includegraphics[width=0.9in, height=1.25in, clip, keepaspectratio]{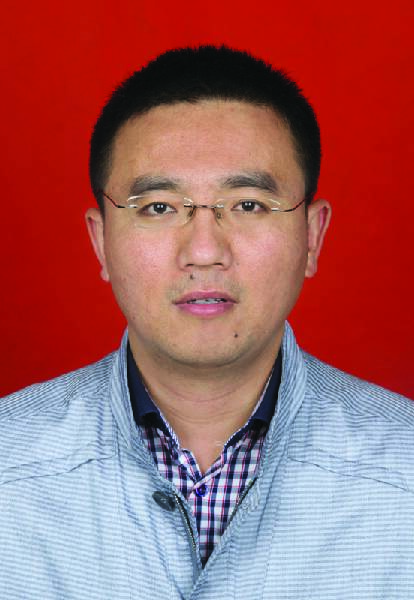}}]{Wei Zhao} received the B.S., M.S. and Ph.D. degrees from Xidian University, Xi’an, China, in 2002, 2005 and 2015, respectively. He is currently a professor in the School of Computer Science and Technology at Xidian University. His research direction is pattern recognition and intelligent systems, with specific interests in attributed graph mining and search, machine learning, signal processing and precision guiding technology.
\end{IEEEbiography}

\begin{IEEEbiography}[{\includegraphics[width=0.9in, height=1.25in, clip, keepaspectratio]{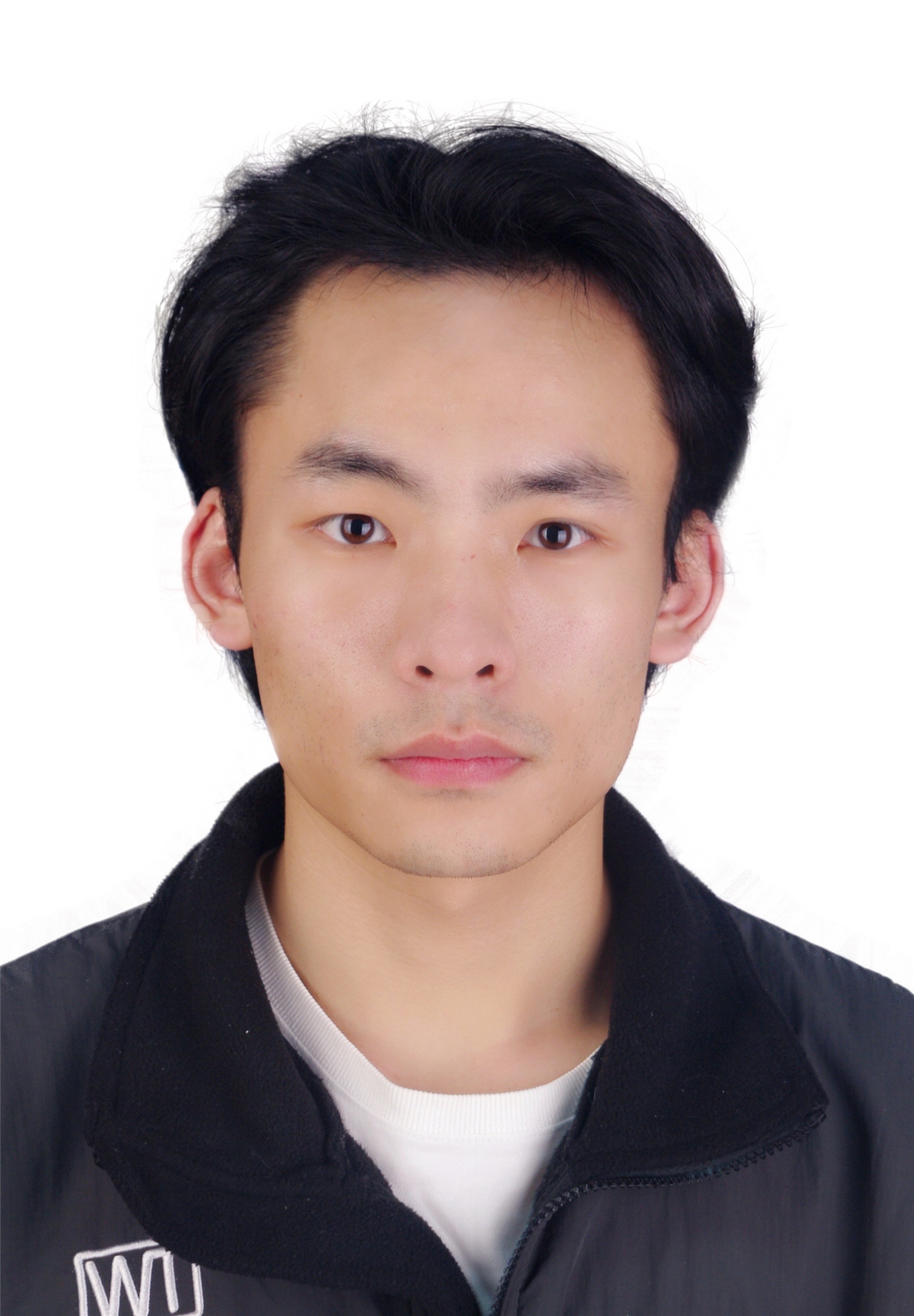}}]{Weigang Lu} received the B.S. degree in Internet of Things from Anhui Polytechnic University, China, in 2019. He is currently working towards a Ph.D. degree with the School of Computer Science and Technology, Xidian University, China. His research interests include data mining and machine learning on graph data.
\end{IEEEbiography}

\begin{IEEEbiography}[{\includegraphics[width=0.9in, height=1.25in, clip, keepaspectratio]{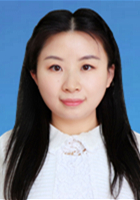}}]{Xinyan Huang} received the B.S. degree from the School of Electronic Science and Technology, Shaanxi University of Science and Technology, Xi'an, China, in 2015, and the M.S. degree from the School of Physics and Information Technology, Shaanxi Normal University, Xi' an, in 2017. She is currently working towards a Ph.D. degree with the School of Artificial Intelligence, Xidian University, China. Her research interests include machine learning, computer vision, and pattern recognition.
\end{IEEEbiography}

\begin{IEEEbiography}[{\includegraphics[width=0.9in, height=1.25in,clip,keepaspectratio]{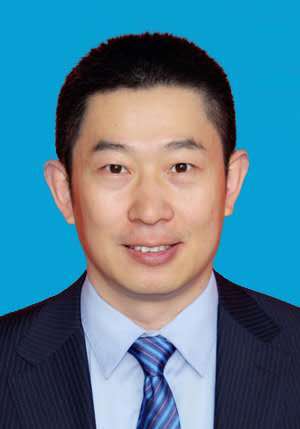}}]{Jiangtao Cui} received the M.S. and Ph.D. degree both in Computer Science from Xidian University, Xian, China in 2001 and 2005 respectively. During 2007 and 2008, he has been with the Data and Knowledge Engineering group working on high-dimensional indexing for large scale image retrieval, in the University of Queensland, Australia. He is currently a professor in the School of Computer Science and Technology, Xidian University. His current research interests include data and knowledge engineering, data security, and high-dimensional indexing. 
\end{IEEEbiography}

\begin{IEEEbiography}[{\includegraphics[width=0.9in, height=1.25in, clip, keepaspectratio]{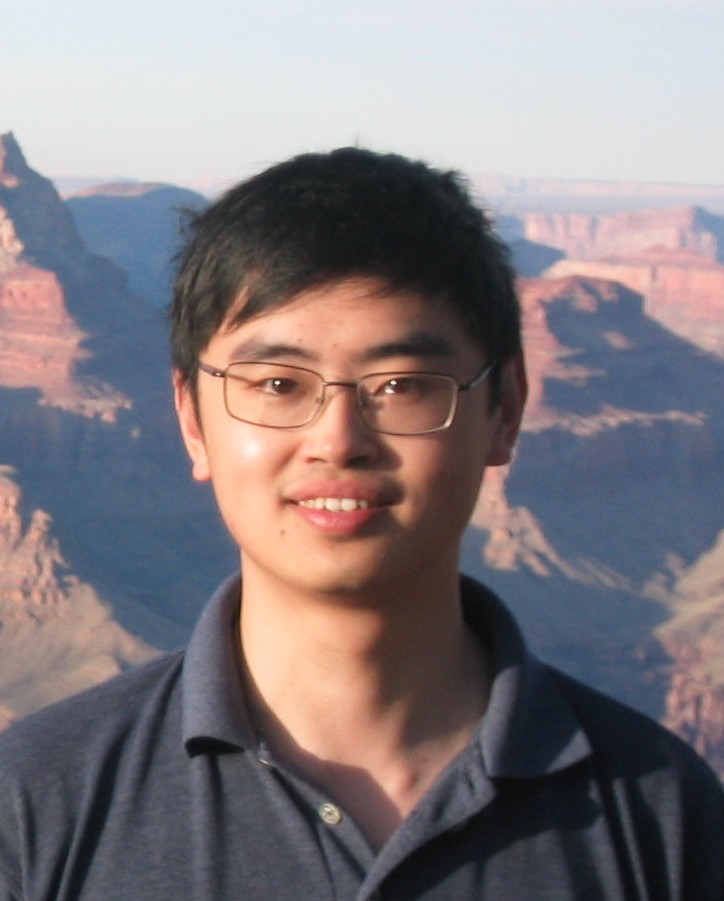}}]{Xiaofei He} received the B.S. degree in computer science from Zhejiang University, China, in 2000, and the Ph.D. degree in computer science from The University of Chicago, in 2005. He is currently a Professor with the State Key Lab of CAD\&CG, Zhejiang University. Prior to joining Zhejiang University, he was a Research Scientist with Yahoo! Research Labs, Burbank, CA, USA. His research interests include machine learning, information retrieval, and computer vision.
\end{IEEEbiography}

\vfill

\end{document}